\documentclass[10pt,twocolumn,letterpaper]{article}

\usepackage{cvpr}      
\usepackage[accsupp]{axessibility}
\usepackage{xspace}
\newcommand{\ours}{\textbf{SmartCLIP}\xspace}

\makeatletter
\@namedef{ver@everyshi.sty}{}
\makeatother
\usepackage{tikz}
\usetikzlibrary{bayesnet,arrows,backgrounds}

\usepackage{tcolorbox}

\usepackage{graphicx}
\usepackage{subcaption}
\renewcommand{\paragraph}[1]{\noindent\textbf{#1}}

\newcommand{\stkout}[1]{\ifmmode\text{\sout{\ensuremath{#1}}}\else\sout{#1}\fi}
\usepackage{amsmath,amsfonts,bm}

\DeclareMathOperator*{\argmin}{arg\,min}

\providecommand{\0}{\mathbf{0}}

\providecommand{\ii}{\mathbf{i}}

\providecommand{\mm}{\mathbf{m}}

\providecommand{\tt}{\mathbf{t}}

\providecommand{\vv}{\mathbf{v}}

\providecommand{\zz}{\mathbf{z}}

\providecommand{\cB}{\mathcal{B}}

\providecommand{\cI}{\mathcal{I}}

\providecommand{\cL}{\mathcal{L}}
\providecommand{\cM}{\mathcal{M}}
\providecommand{\cN}{\mathcal{N}}

\providecommand{\cT}{\mathcal{T}}

\providecommand{\cV}{\mathcal{V}}

\providecommand{\cZ}{\mathcal{Z}}

\providecommand{\R}{\mathbb{R}} 

\providecommand{\mI}{\mathbf{I}}

\providecommand{\mP}{\mathbf{P}}

\providecommand{\mT}{\mathbf{T}}

\newcommand{\norm}[1]{\left\lVert#1\right\rVert}

\usepackage[noend]{algpseudocode}

\errorcontextlines\maxdimen

\makeatletter
    \newcommand*{\algrule}[1][\algorithmicindent]{\makebox[#1][l]{\hspace*{.5em}\thealgruleextra\vrule height \thealgruleheight depth \thealgruledepth}}%
\newcommand*{\thealgruleextra}{}
\newcommand*{\thealgruleheight}{.75\baselineskip}
\newcommand*{\thealgruledepth}{.25\baselineskip}

\newcount\ALG@printindent@tempcnta
\def\ALG@printindent{%
    \ifnum \theALG@nested>0
        \ifx\ALG@text\ALG@x@notext
        \else
            \unskip
            \addvspace{-1pt}
            \ALG@printindent@tempcnta=1
            \loop
                \algrule[\csname ALG@ind@\the\ALG@printindent@tempcnta\endcsname]%
                \advance \ALG@printindent@tempcnta 1
            \ifnum \ALG@printindent@tempcnta<\numexpr\theALG@nested+1\relax
            \repeat
        \fi
    \fi
    }%
\usepackage{etoolbox}
\patchcmd{\ALG@doentity}{\noindent\hskip\ALG@tlm}{\ALG@printindent}{}{\errmessage{failed to patch}}
\makeatother

\newbox\statebox
\newcommand{\myState}[1]{%
    \setbox\statebox=\vbox{#1}%
    \edef\thealgruleheight{\dimexpr \the\ht\statebox+1pt\relax}%
    \edef\thealgruledepth{\dimexpr \the\dp\statebox+1pt\relax}%
    \ifdim\thealgruleheight<.75\baselineskip
        \def\thealgruleheight{\dimexpr .75\baselineskip+1pt\relax}%
    \fi
    \ifdim\thealgruledepth<.25\baselineskip
        \def\thealgruledepth{\dimexpr .25\baselineskip+1pt\relax}%
    \fi
    \State #1%
    \def\thealgruleheight{\dimexpr .75\baselineskip+1pt\relax}%
    \def\thealgruledepth{\dimexpr .25\baselineskip+1pt\relax}%
}

\usepackage{amsmath,amssymb,amsthm}
\usepackage{thmtools,thm-restate}
\usepackage{mathtools}

\theoremstyle{plain}
\newtheorem{theorem}{Theorem}[section]

\theoremstyle{definition}
\newtheorem{definition}[theorem]{Definition}
\theoremstyle{remark}

\usepackage{enumitem}
\usepackage{soul}
\usepackage{multirow}
\usepackage{algorithm}
\usepackage{xcolor}
\usepackage{caption}
\usepackage{wrapfig}

\providecommand{\vi}{\ii}
\providecommand{\vt}{\mathbf{t}}

\providecommand{\ei}{\bm\epsilon_{\mathrm{I}}}
\providecommand{\et}{\bm\epsilon_{\mathrm{T}}}

\providecommand{\zi}{\zz_{\mathrm{I}}}
\providecommand{\zt}{\zz_{\mathrm{T}}}
\providecommand{\tzi}{\tilde{\zz}_{\mathrm{I}}}

\providecommand{\gii}{g_{\mathrm{I}}}
\providecommand{\gtt}{g_{\mathrm{T}}}
\providecommand{\fii}{f_{\mathrm{I}}}
\providecommand{\ftt}{f_{\mathrm{T}}}
\providecommand{\htt}{h_{\mathrm{T}}}
\providecommand{\hii}{h_{\mathrm{I}}}
\providecommand{\rtt}{r_{\mathrm{T}}}
\providecommand{\rii}{r_{\mathrm{I}}}

\providecommand{\hzi}{\hat{\zz}_{\mathrm{I}}}
\providecommand{\hzt}{\hat{\zz}_{\mathrm{T}}}
\providecommand{\cVi}{\cI}
\providecommand{\cVt}{\cT}

\providecommand{\cZi}{\cZ_{\mathrm{I}}}

\providecommand{\Ls}{L_{\mathrm{sparsity}} }
\providecommand{\La}{L_{\mathrm{align}} }

\definecolor{cvprblue}{rgb}{0.21,0.49,0.74}
\usepackage[pagebackref,breaklinks,colorlinks,allcolors=cvprblue]{hyperref}
\definecolor{cocoColor}{HTML}{FFCCCC}
\definecolor{llavaColor}{HTML}{FFE4E1}
\definecolor{sharegptColor}{HTML}{E0FFFF}
\definecolor{pinkbox}{RGB}{255,192,203}
\definecolor{orangebox}{RGB}{255,218,185}
\definecolor{bluebox}{RGB}{173,216,230}
\definecolor{bluetext}{RGB}{0,0,255}
\usepackage{multirow}
\usepackage{multicol}
\usepackage{hyperref}
\usepackage{array}
\usepackage{adjustbox}
\definecolor{pastelgreen}{RGB}{188, 212, 187}
\definecolor{darkgreen}{RGB}{0,100,0}
\usepackage{xcolor}
\definecolor{softblue}{RGB}{70, 136, 207}        
\definecolor{dustyrose}{RGB}{220, 140, 156}      
\definecolor{sageleaf}{RGB}{136, 176, 75}        
\definecolor{goldenyellow}{RGB}{255, 198, 109}   
\definecolor{lavender}{RGB}{150, 123, 182}       
\definecolor{seafoam}{RGB}{98, 182, 177}         
\definecolor{terracotta}{RGB}{226, 114, 91}      
\definecolor{periwinkle}{RGB}{143, 161, 215}     
\definecolor{mintgreen}{RGB}{152, 203, 178}      
\definecolor{mauve}{RGB}{224, 176, 187}          

\def\paperID{6793} 
\def\confName{CVPR}
\def\confYear{2025}

\title{SmartCLIP: Modular Vision-language Alignment with Identification Guarantees}

\author{Shaoan Xie*$^{1,2}$, Lingjing Kong*$^{1}$, Yujia Zheng$^{1}$, Yu Yao$^{3}$, 
Zeyu Tang$^{1}$, \\ Eric P. Xing$^{1,2}$, Guangyi Chen$^{2, 1}$, Kun Zhang$^{1,2}$\\
\tt\small $^1$ Carnegie Mellon University \\
\tt\small $^2$ Mohamed bin Zayed University of Artificial Intelligence\\
\tt\small $^3$ The University of Sydney\\
{\footnotesize * Equal contribution}
\\
}

\begin{document}
\maketitle
\begin{abstract}
Contrastive Language-Image Pre-training (CLIP)~\citep{radford2021learning} has emerged as a pivotal model in computer vision and multimodal learning, achieving state-of-the-art performance at aligning visual and textual representations through contrastive learning.
However, CLIP struggles with potential information misalignment in many image-text datasets and suffers from entangled representation. 
On the one hand, short captions for a single image in datasets like MSCOCO may describe disjoint regions in the image, leaving the model uncertain about which visual features to retain or disregard.
On the other hand, directly aligning long captions with images can lead to the retention of entangled details, preventing the model from learning disentangled, atomic concepts -- ultimately limiting its generalization on certain downstream tasks involving short prompts.

In this paper, we establish theoretical conditions that enable flexible alignment between textual and visual representations across varying levels of granularity. 
Specifically, our framework ensures that a model can not only \emph{preserve} cross-modal semantic information in its entirety but also \emph{disentangle} visual representations to capture fine-grained textual concepts. 
Building on this foundation, we introduce \ours, a novel approach that identifies and aligns the most relevant visual and textual representations in a modular manner. 
Superior performance across various tasks demonstrates its capability to handle information misalignment and supports our identification theory. The code is available at \url{https://github.com/Mid-Push/SmartCLIP}.
\end{abstract}
\section{Introduction} \label{sec:introduction}
Contrastive Language-Image Pre-training (CLIP)~\citep{radford2021learning} has been the cornerstone for many computer vision and machine learning tasks, such as text-to-image retrieval~\citep{baldrati2023zero}, image and video understanding~\citep{che2023enhancing,zhang2022pointclip,wang2023exploring,tang2021clip4caption,ju2022prompting,rasheed2023fine}, and generative models~\citep{ramesh2022hierarchical,rombach2022high,liu2024visual}. It aligns the representations from different modalities with a contrastive learning loss~\citep{oord2018representation,chen2020simple}. 
Specifically, each image-caption pair in the dataset is treated as a positive pair, while negative pairs are created by matching images with captions randomly drawn from the dataset.
The image and text encoders are trained with a symmetric cross-entropy loss that draws the image and text representations in each positive pair together while pulling the negative pairs' representations apart.

\begin{figure}
    \centering
    \includegraphics[width=1\linewidth]{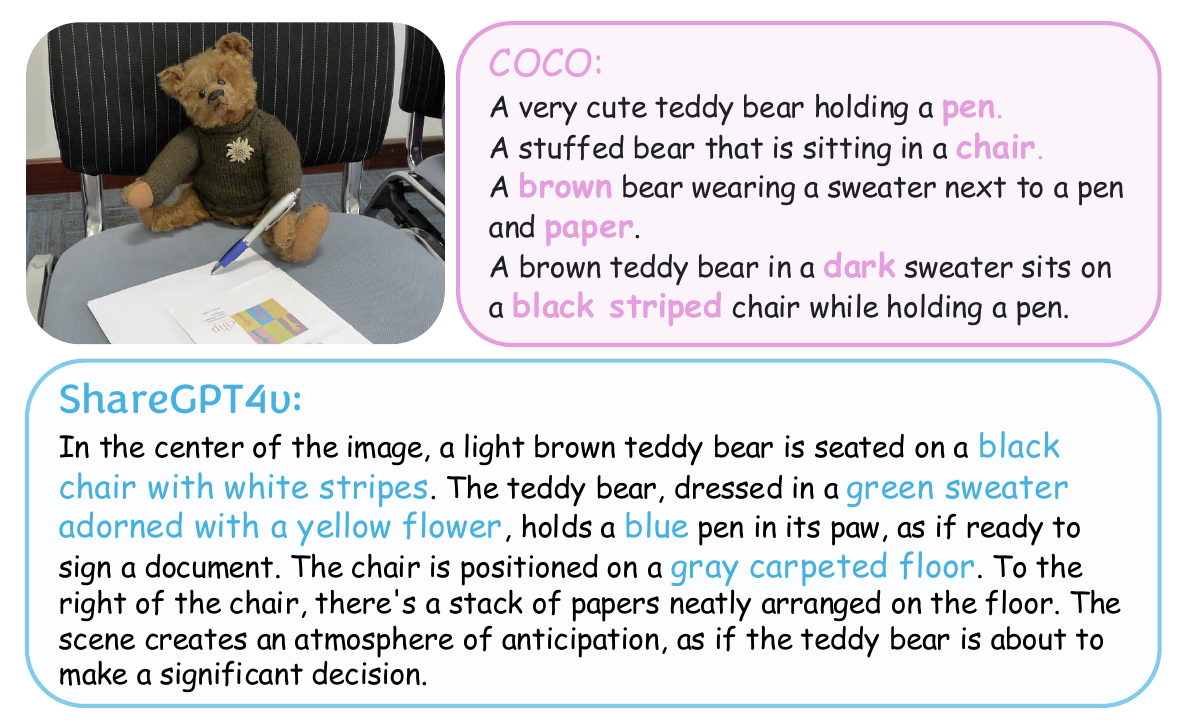}
    \caption{
        \textbf{Depiction of two primary challenges for CLIP.} 
        (1) Information Misalignment: An image can be paired with multiple captions that describe disparate aspects - the first caption contains concepts ``bear'' and ``pen'' whereas the second only mentions ``bear'' and ``paper''.
        Aligning the image with both captions leads to the loss of key concepts ``pen'' and ``paper'' not shared across the captions.
        (2) Entangled Representations: Long, detailed captions involving multiple concepts (e.g., ``chair'', ``pen'', ``flower'', ``floor'') encourage the model to form entangled representations, hindering independent understanding of each individual concept.
    }
    \label{fig:illus}
    \vspace{-0.5cm}
\end{figure}

Training CLIP requires vast amounts of image-text pairs, making it challenging to maintain dataset quality at such a large scale. 
In particular, the quality of text captions has been a key concern, prompting the development of various methods to enhance both their diversity and accuracy.
ALIGN~\citep{jia2021scaling} shows that scaling up the dataset size can compensate for the noisy text supervision.
BLIP models~\citep{li2022blip,li2023blip} improve captions by incorporating additional captioning and filtering mechanisms.
VE-CLIP \citep{lai2024veclip} introduces a visual-enriched captioning approach to further refine caption quality.
Similarly, LaCLIP~\citep{fan2024improving} leverages language models to rewrite captions, while RecapCLIP~\citep{li2024if} uses LLaMA-3~\citep{meta2024introducing} to generate captions for 1.3 billion images.
Despite these efforts, recent findings reveal that longer and seemingly higher-quality captions do not necessarily yield improved performance on many downstream tasks~\citep{lai2024revisit}.
\citet{li2024if} find that the text-to-image retrieval accuracy on Flickr30K drops from 84.2 to 74.1 when replacing the original captions with longer captions. 

A key issue contributing to the observed performance degradation is the information misalignment between images and their captions, a problem that becomes more pronounced when multiple captions are paired with a single image. 
On one hand, an image may be paired with several captions, each capturing only a partial aspect of the image. In Figure~\ref{fig:illus}, aligning the image with the first caption, ``A very cute teddy bear holding a pen'', risks forcing the model to discard other important concepts like ``chair'' and ``paper'', which are required to align with the second and third captions in the pink text box. This misalignment between image and text introduces conflicts during standard CLIP training, leading to the loss of key visual concepts.

On the other hand, training CLIP with long, detailed captions, as seen in recent approaches~\citep{lai2024veclip,fan2024improving,li2024if}, encourages the model to learn entangled representations of multiple concepts bundled together in a single caption.
Thus, it remains challenging to explicitly extract object/concept-centric representations from CLIP’s visual representation.
This entanglement is particularly problematic for tasks that require individual, atomic concepts or novel combinations of them, as empirically observed on short-text-to-image retrieval tasks~\citep{zhang2024long}.
In Figure~\ref{fig:illus}, the long caption generated by ShareGPT4V~\citep{chen2023sharegpt4v} contains an exhaustive set of concepts such as ``chair'', ``pen'', ``flower'', and ``floor''. 
This aggregation can hinder the model’s performance on tasks that demand the understanding of each concept individually.

In this paper, we propose a refined approach to representation alignment in vision-language models like CLIP~\citep{radford2021learning}. 
We frame the alignment challenge as a latent-variable identification problem and develop theoretical conditions that enable flexible alignment between textual and visual representations at different levels of granularity. 
Our framework enables the model to \emph{preserve} the complete cross-modal information while also \emph{disentangling} representations to capture fine-grained concepts, effectively addressing the misalignment and disentanglement issues discussed earlier.

Building on these theoretical insights, we introduce \ours, a novel method that identifies and aligns visual and textual concepts in a modular manner.
Specifically, we design a mask network that selects a subset of dimensions from the full representations, corresponding to only the concepts present in each specific caption. 
This allows the model to perform text-image alignment over the most relevant concepts modules, rather than the entire representation.
We empirically demonstrate that \ours outperforms state-of-the-art models across a range of downstream tasks, showcasing its effectiveness in addressing alignment challenges. 
In particular, \ours significantly improves retrieval performance across text lengths, achieving 98.7\% accuracy (up from 78.2\%) on the ShareGPT4V long text-to-image retrieval tasks, while boosting short text-to-image retrieval R1 from 56.1\% to 66.0\%.
\\
\noindent Our main contributions are summarized as follows.
\begin{enumerate}[label=\roman*, leftmargin=1em, topsep=0.5pt, partopsep=0pt, itemsep=-0.0em] 
    \item We identify critical issues of information misalignment and entangled representations within the CLIP framework. To overcome these challenges, we propose a latent-variable formulation and establish theoretical conditions that guarantee the recovery of the latent variables.
    \item Building upon our theoretical findings, we propose \ours, featuring adaptive masking and a modular contrastive learning objective that facilitates the learning of disentangled, modular representations.
    \item We perform extensive experiments on a variety of tasks, including long and short text-to-image retrieval, zero-shot classification, and text-to-image generation. \ours consistently outperforms or matches state-of-the-art models across these benchmarks, demonstrating its efficacy and validating our theoretical contributions.
\end{enumerate}

\section{Related Work}

\paragraph{Vision-language models.}
The breakthrough of CLIP~\citep{radford2021learning} has attracted significant attention from the community.
SLIP \citep{mu2022slip} and DECLIP \citep{li2021supervision} propose to incorporate self-supervised learning techniques to improve the learned representation. 
Coca \citep{yu2022coca} introduces a decoder in addition to the contrastive learning branch. 
LiT~\citep{zhai2022lit} locks the image encoder and only finetunes the text encoder. SigLIP \citep{zhai2023sigmoid} adopts a simple sigmoid loss to handle large training batch sizes. 
LoTLIP~\citep{wu2024lotlip} inserts corner tokens after the classification token to support long-text understanding. TULIP~\citep{najdenkoska2024tulip} replaces the absolute position embedding with the relative position embedding to support longer text understanding. 
ALIGN~\citep{jia2021scaling} demonstrates that increasing dataset size can mitigate the impact of noisy text supervision. 
Recent methods have been focusing on generating better captions ~\citep{li2022blip,li2023blip,lai2024revisit,lai2024veclip,li2024if,zheng2024dreamlip}.
CLIP-MOE \cite{zhang2024clip} introduces mixture-of-experts to CLIP. LLM2CLIP \cite{huang2024llm2clip} augments CLIP with large language models.
LongCLIP~\citep{zhang2024long} extends the token constraint of CLIP from 77 to 248 and applies PCA to perform short text-to-image contrastive learning to preserve its short text capability.
Llip~\citep{lavoie2024modeling} learns a text-dependent visual representation by mixing a set of learnable tokens with a cross-attention module.
In contrast, \ours directly learns a single global representation that encodes all disentangled, interpretable concepts through masking.

\paragraph{Latent variable identification.}
Learning high-level, semantic information from low-level observational data (e.g., images and text) can often be formulated as latent-variable identification problems.
Though appealing, such tasks are accompanied by substantial difficulties, especially for complex real-world data distributions involving nonlinear generating functions.
Recently, a wealth of papers~\citep{hyvarinen2019nonlinear,khemakhem2020variational,zhang2024causal,buchholz2024learning,von2024nonparametric,zhang2024identifiability,ahuja2023interventional,yaomulti,sturma2023unpaired,morioka2023connectivity,morioka2024causal,daunhaweridentifiability,gresele2020incomplete} propose to overcome such obstacles by leveraging auxiliary information, such as temporal information, multiple domains, and multiple views/modalities.
Especially related to our work are those that tap into the paired multi-view data to identify the shared information across available views~\citep{morioka2024causal,gresele2020incomplete,morioka2023connectivity,yaomulti,von2021self,daunhaweridentifiability}.
Recent work~\citep{morioka2024causal,gresele2020incomplete,morioka2023connectivity} relies on specific forms of latent variable distributions (e.g., independence or exponential family).
These constraints restrict their applicability for distributions entailing complex interactions among latent variables.
Prior work~\citep{von2021self,daunhaweridentifiability} adopt more flexible assumptions on the underlying distribution and identify blocks of latent variables directly shared by two views arising from data augmentations, which is extended to a multi-view setting~\citep{yaomulti}.
The problem under investigation can be viewed as a form of this multi-view setting where the paired image and text captions are considered as views sharing the semantic latent variable.
Existing works~\citep{von2021self,daunhaweridentifiability,yaomulti} assume that views are grouped over all the data pairs and this grouping information is known so that one can learn a designated encoder for each view group.
However, this view grouping information is inaccessible for our problems -- for any two text captions of different images, we cannot judge whether they belong to the same view group.
In our theory section, we show that by properly utilizing the data-generating process, we can learn such information directly and further achieve the desired identification results, thus generalizing existing multi-view latent variable identification results.






\section{Problem Formulation}

As motivated previously, we aim to 1) preserve all the semantic information shared across modalities, and 2) learn a disentangled representation that corresponds to textual concepts at diverse granularity levels.
To this end, we propose the following data-generating process underlying the visual-language data distribution.

\paragraph{Notations.}
We indicate the dimensionality of a vector with $ d(\cdot) $.
We denote a subset of dimensions of a vector $\zz$ with $ [\zz]_{\cB} $ with the index set $\cB$.  
We define the set of indices whose corresponding values are nonzero in a vector $\mm$ with $ \cB(\mm):= \{ i \in d(\mm): [\mm]_{i} \neq 0 \} $.

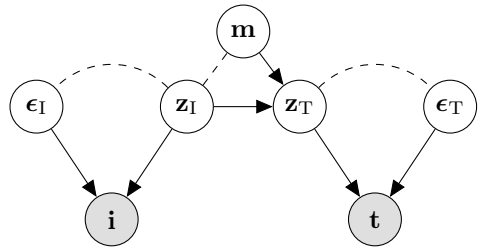
\begin{figure}[ht]
    \centering
    \begin{tikzpicture}
        \node[obs] (v1) at (1, 0) {$\vi$};
        \node[obs] (v2) at (4.5, 0) {$\vt$};
        \node[latent] (s1) at (0, 1.5) {$\ei$};
        \node[latent] (s2) at (5.5, 1.5) {$\et$};
        \node[latent] (z) at (2, 1.5) {$\zi$};
        \node[latent] (tz) at (3.5, 1.5) {$\zt$};
        \node[latent] (m) at (2.75, 2.5) {$\mm$};
        
        \draw[-,dashed] (z) to[bend left=315] (s1);  
        \draw[-,dashed] (tz) to[bend left=45] (s2);  
        \draw[-,dashed] (z)to(m);
        \edge {s1} {v1};
        \edge {s2} {v2};
        \edge {z} {v1,tz};
        \edge {tz} {v2};
        \edge {m} {tz};
    \end{tikzpicture}
    \caption{
        \footnotesize
        \textbf{The data-generating process.}
        The text representation $\zt$ contains partial information of the vision representation $\zi$, as indicated by masking $\mm$.
        Modal-specific information is represented as $\ei$ and $\et$.
        Dashed edges indicate potential statistical dependence.
    }
    \label{fig:causal_model}
\end{figure}

\paragraph{Data-generating processes.}
We depict the data-generating process in Figure~\ref{fig:causal_model} and in \eqref{eq:data_generating_process}.
\begin{align} \label{eq:data_generating_process}
    \zt := \zi \odot \mm; \, \vi := \gii (\zi, \ei); \, \vt := \gtt(\zt, \et). 
\end{align}
We assume that each pair of image $\vi \in \cVi \subset \R^{ d(\vi) }$ and text caption $\vt \in \cVt \subset \R^{ d(\vt) } $ originate from semantic information $ \zi \in \cZi \subset \R^{d(\zi)}$, together with modality-specific variations $ \ei $
and $ \et $ (e.g., illumination for images, tenses for text), through generating functions $ \gii: (\zi, \ei) \mapsto \vi $ and $ \gtt: (\zt, \et) \mapsto \vt $ respectively.
We treat the text caption $\tt$ as continuous variables, as each word can be represented with a continuous word embedding vector in practice~\citep{bengio2000neural,mikolov2013distributed}.

As demonstrated in Figure~\ref{fig:illus}, text captions of the same image often convey partial information of the entire image semantics. 
Thus, we associate each text caption's representation $\zt:= \mm \odot \zi$ with a binary random mask $\mm \in \cM \subset \{0, 1\}^{d (\zi)}$ that eliminates information absent in the specific caption $\vt$.

\paragraph{Goals.}
Our two goals can be formalized as follows.
\begin{enumerate}[label=\alph*,leftmargin=1em, topsep=0.5pt, partopsep=0pt, itemsep=-0.0em]
    \item \label{goal:extract_zi} \textit{Preserving cross-modal information}: identifying the complete latent representation $\zi$.
    \item \label{goal:disentangle_zi} \textit{Disentangling concepts}: identifying concepts $ \zt $ associated with a given textual description $ \vt $ at different granularity levels potentially unseen during training.
\end{enumerate}

\paragraph{Examples.}
In Figure~\ref{fig:illus}, the image $\vi$ contains concepts ``bear'', ``chair'', and ``pen'', which we assume correspond to three components in the representation $\zi$, say $ [\zi]_{1} $, $[\zi]_{2}$, and $ [\zi]_{3}$.
The first two COCO captions $\vt^{(1)}$ and $\vt^{(2)}$ mention only a subset of these concepts, i.e., (``bear'', ``pen'') and (``bear'', ``chair'') separately.
Thus, the masks for these captions are $ \mm^{(1)} = [1, 0 , 1] $ and $ \mm^{(2)} = [1, 1, 0] $.
The variables $\ei$ and $\et$ represent modality-specific nuance factors such as illumination conditions in the image $\vi$ and syntax in the text $\vt$.
For Goal~\ref{goal:extract_zi}, we seek to \textit{preserve} the complete information $\zi$.
This involves retaining all relevant textual concepts present in the captions, namely ``bear'', ``chair'', and ``pen'' from captions $\vt^{(1)}$ and $\vt^{(2)}$.
For Goal~\ref{goal:disentangle_zi}, we intend to \textit{disentangle} the representation $\zi$ into finer concept blocks potentially unseen in the training.
This includes identifying individual concepts such as ``bear'' in the dimension $[\zi]_{1}$, even if the training captions only include ``bear'' in combination with other concepts.

\section{Identification Theory} \label{sec:theory}

In this section, we present the theoretical results towards Goal~\ref{goal:extract_zi} and Goal~\ref{goal:disentangle_zi}.
We show that under a suitable learning objective \eqref{eq:contrastive_objective}, the learned variables $ ( \hzi, \hzt) $ can be identified with the corresponding true quantities $(\zi, \zt)$ up to certain equivalent classes.
In particular, we resort to the block-wise identifiability~\citep{von2021self,lachapelle2023additive,yaomulti,kong2022partial} throughout this work.
This suffices for our purpose since often several dimensions jointly (i.e., a block) comprise a meaningful concept while a single dimension may not be interpretable.
\vspace{0.4em}
\begin{definition}[Block-wise Identifiability] \label{def:identifiability}
    The true variable $\vv$ is block-wise identifiable if it is related to its estimate $\hat{\vv}$ through an invertible map $ \vv \mapsto \hat{\vv} $.
\end{definition}


\paragraph{The learning objective.}
Our estimation model consists of vision/text encoders $(\fii, \ftt)$ (smooth, invertible functions), and a masking function $ \hat{\mm}: \cVt \to \cM $ that estimates the true mask $\mm$ underlying a given text caption $\vt$.  
\begin{align} \label{eq:contrastive_objective}
    \begin{split}            
        &\underbrace{\argmin_{ \fii, \ftt, \hat{\mm} } \norm{ \hat{\mm} (\vt) }_{0}}_{ \Ls }, \quad \text{subject to: } \\
        & \underbrace{\argmin_{ \fii, \ftt, \hat{\mm} } \norm{ \fii (\vi) \odot \hat{\mm} (\vt) - \ftt (\vt) }}_{ \La }, \, \forall (\vi, \vt).
    \end{split}
\end{align}
Our learning objective \eqref{eq:contrastive_objective} consists of an alignment term $ \La $ that draws the positive pairs across modalities.
The negative pairs in regular contrastive losses~\citep{radford2021learning,oord2018representation,chen2020simple} can be implemented through an entropy term at the sample limit~\citep{wang2020understanding}.
This serves the same role as the invertibility condition on the encoder models~\citep{von2021self}, which we directly assume for theoretical convenience.
In Section~\ref{sec:method}, we discuss practical considerations for constructing negative pairs.
We enforce sparsity regularization $\Ls$ on the inferred mask $\hat{\mm}$ to select the simplest representation.

We introduce our key conditions in Condition~\ref{cond:identification_conditions} and theoretical results in Theorem~\ref{thm:identification}.
\begin{restatable}[Identification Conditions]{condition}{identificationconditions} \label{cond:identification_conditions} {\ }
    \begin{enumerate}[label=\textit{\roman*},leftmargin=2em, topsep=0.5pt, partopsep=0pt, itemsep=-0.0em]
    \setlength\itemsep{-0.0em}
        \item \label{asmp:invertibility} [Smoothness \& invertibility]: Generating functions $\gii$ and $\gtt$ are smooth and have smooth inverses.
    
        \item \label{asmp:full_support} [Fully-supported $p(\zi, \mm)$]: The joint distribution over the semantic variable $\zi$ and the mask $\mm$ is fully supported: $ p(\zi, \mm) > 0 $ for any $ (\zi, \mm) \in \cZi \times \cM $.
    \end{enumerate}
\end{restatable}

\paragraph{Discussion.}
Condition~\ref{cond:identification_conditions}-\ref{asmp:invertibility} ensures the generating functions $(\gii, \gtt)$ preserve latent variables' information, without which the task of recovering such latent variables would be ill-posed.
Practically, the high dimensionality of image data $\vi$ offers sufficient capacity to hold all information, and the text variable $\vt$ only contains information filtered through its mask $\mm$.
This condition is widely employed in the latent-variable identification literature~\citep{hyvarinen2016unsupervised,khemakhem2020variational,von2021self,kong2022partial}.
Condition~\ref{cond:identification_conditions}-\ref{asmp:full_support} prescribes that the representation $ \zi $ and the mask $ \mm $ that marginally appear in the training distribution should also be present jointly with non-zero probability density.
Interpreting the mask $\mm$ as a concept selector (e.g., selecting "bear" and "pen"), this condition ensures that each concept retains its full range of variations (such as different shapes of bears and lengths of pens) across various mask selections. 
To satisfy this requirement, one can restrict the joint support $\cZi \times \cM$ to an appropriate subset, ensuring that only relevant combinations of $\zi$ and $\mm$ are present.
Alternatively, one can enrich the caption set for each image, thereby increasing the diversity and coverage of concept combinations and filling in the joint support.
This aligns with recent caption-augmentation techniques~\citep{li2024if,li2022blip,li2023blip,lai2024veclip,zhang2024long} as discussed in Section~\ref{sec:introduction}, revealing the synergy between our framework and existing efforts in the community.

\begin{restatable}[Concept Representation Identification]{theorem}{identificationtheory} \label{thm:identification} {\ }
    We assume the data-generating process in \eqref{eq:data_generating_process}.
    Let $ (\fii, \ftt, \hat{\mm} )$ be an optimum of \eqref{eq:contrastive_objective}.
    Under Condition~\ref{cond:identification_conditions}, the true representation $[\zz]_{\tilde{\cB}}$ is block-wise identifiable for any index set $\tilde{\cB}$ such that $ \tilde{\cB} = \cup_{ \mm \in \cV } \cB(\mm) $ or $ \tilde{\cB} = \cap_{ \mm \in \cV } \cB(\mm) $ over any subset of masks $ \cV \subset \cM $.
\end{restatable}

\paragraph{Concept preservation.}
Theorem~\ref{thm:identification} states that one can recover the concept block $ [\zi]_{\cB(\mm)} $ associated with each individual text caption in the dataset $ \cM $.~\footnote{
    We refer to a text caption with its mask $ \mm $ to simplify the notation.
}
Furthermore, it ensures that the union of concepts $[\zi]_{\cup_{\mm \in \cV} \cB(\mm)}$ from any subset of text captions $\cV \subset \cM$ can be preserved.
In the running example of Figure~\ref{fig:illus}, our formulation allows us to preserve concepts (``bear'', ``pen'', ``chair'') in the image representation by selectively matching them with the two captions, whereas existing models like CLIP may lose either ``pen'' and ``chair'' since they are only mentioned in one caption.
Therefore, Theorem~\ref{thm:identification} effectively addresses Goal~\ref{goal:extract_zi}.

\paragraph{Concept disentanglement.}
The intersection operation in Theorem~\ref{thm:identification} empowers us to \textit{disentangle} representations into potentially atomic concepts.
In the example of Figure~\ref{fig:illus}, we can identify the concept ``bear'' as the intersection of the two text captions, despite the absence of a standalone caption containing only ``bear'' in the dataset.
Consequently, this part of the statement tackles Goal~\ref{goal:disentangle_zi}.
Our results underscore the importance of associating each image with a diverse set of captions that share overlapping concepts.


\paragraph{Theoretical contribution.}
Theorem~\ref{thm:identification} extends existing theoretical frameworks~\citep{von2021self,yaomulti,daunhaweridentifiability}.
Notably, \citet{yaomulti} provide identification guarantees for shared representations over multiple views, generalizing earlier results confined to two views~\citep{von2021self,daunhaweridentifiability}.
This multi-view formulation is analogous to our setup when considering each group of text captions $\vt$ associated with the same mask $\mm$ as a distinct view group.
However, prior work~\citep{yaomulti} relies on explicit knowledge of these groupings to train view-specific encoders.
In contrast, our problem setting presents a greater challenge since we have no access to this grouping information. 
Specifically, given two captions of any different images, it is unclear whether they stem from the same mask (i.e., the same view group).
Therefore, the identification guarantees in prior studies do not apply to our setting.
Theorem~\ref{thm:identification} demonstrates that our estimation model, paired with the learning objective \eqref{eq:contrastive_objective}, can automatically infer the necessary grouping information (i.e., the masks). By doing so, our approach relaxes the identification conditions in previous work, enabling effective representation identification without explicit group knowledge.

\begin{figure*}[ht]
    \centering
    \includegraphics[scale=0.35]{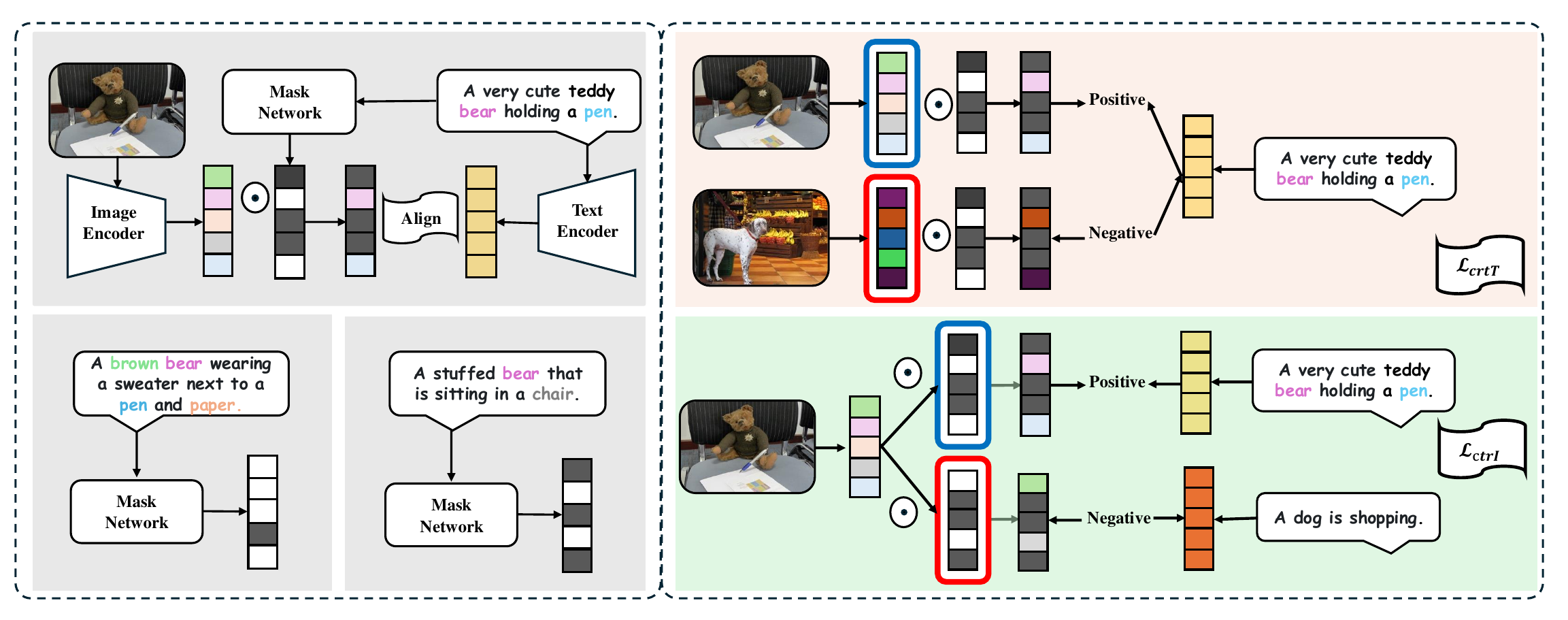}
    \caption{
       \textbf{The diagram of \ours}. On the left, we introduce adaptive masking for alignment with different text prompts. The mask network selects which part of the image representation to be used. On the right, we present our modular contrastive objectives \eqref{eq:ours_symmetric_contrast}.
    }
    \label{fig:model_arch}
\end{figure*}

\vspace{-0.2cm}
\section{\ours: Modular Vision-language Alignment} \label{sec:method}
\vspace{-0.0cm}
Drawing on the theoretical framework in Section~\ref{sec:theory}, we present \ours, a modular alignment model designed to achieve Goal~\ref{goal:extract_zi} and Goal~\ref{goal:disentangle_zi}.
We discuss the implementation of the learning objective \eqref{eq:contrastive_objective} and the model architecture.

\paragraph{Modular alignment through adaptive masking.}
The masking function $\hat{\mm} (\cdot)$ is instrumental in our modular alignment objective~\eqref{eq:contrastive_objective}.
\ours implements this function with a transformer block that ingests a caption representation $ \hzt $ as its input and outputs a binary vector $ \hat{\mm} (\hzt) $ via a straight-through estimator~\citep{bengio2013estimating}.

\paragraph{Modular contrast construction.}
As discussed in Section~\ref{sec:theory}, the negative pairs in regular contrastive losses~\citep{oord2018representation,chen2020simple} serves a similar role as the invertibility assumption (Condition~\ref{cond:identification_conditions}-\ref{asmp:invertibility})~\citep{von2021self,wang2020understanding}.
We denote generic image, text representations as $ \mI $, $\mT$, and positive, negative pairs with $\mP_{\mathrm{pos}}$, $\mP_{\mathrm{neg}}$ respectively.
The canonical one-side contrastive loss $ \cL_{\mathrm{ctr}} $~\cite{radford2021learning} is defined as:
\begin{align} \label{eq:one_side_contrast}
    \begin{split}
        &\mathcal{L}_{\mathrm{ctr}} \bigg( \underbrace{\left( \mI^{(i)}, \mT^{(i)} \right)}_{\mP_{\mathrm{pos}}}, 
        \underbrace{ \left(\mI^{(i)}, \mT^{(j)} \right) }_{ \mP_{\mathrm{neg}} } \bigg) \\
        =& -\frac{1}{N} \sum_{i=1}^{N} \log \frac{\exp\left(\frac{\text{sim}(\mI^{(i)}, \mT^{(i)})}{\tau}\right)}{\sum_{j=1}^{N} \exp\left(\frac{\text{sim}(\mI^{(i)}, \mT^{(j)})}{\tau}\right)},
\end{split}
\end{align}
where we denote the sample size, temperature, and cosine similarity with $ N $, $ \tau $, and $ \text{sim}(\cdot) $ respectively.


Following the symmetric contrastive loss for CLIP~\citep{radford2021learning}, our alignment loss consists of two contrastive loss terms $ \cL_{ \mathrm{ctrI} } $ and $ \cL_{ \mathrm{ctrT} } $ that differ in the negative pairs:
\begin{align} \label{eq:ours_symmetric_contrast}
    \cL_{ \mathrm{ctrI} } := \cL_{ \mathrm{ctr} } \left( \mP_{\mathrm{pos}}, \mP_{\mathrm{negI}} \right), \;
    \cL_{ \mathrm{ctrT} } := \cL_{ \mathrm{ctr} } \left( \mP_{\mathrm{pos}}, \mP_{\mathrm{negT}} \right),
\end{align}
with the positive and negative pairs defined as follows:
\begin{align}
    \mP_{\mathrm{pos}} &:= \left( \hzi^{(i)} \odot \hat{\mm} ( \hzt^{(i)} ), \; \hzt^{(i)} \right), \\
    \mP_{\mathrm{negI}} &:= \left( \hzi^{(i)} \odot \hat{\mm} ( \hzt^{(j)} ), \; \hzt^{(j)} \right), \\
    \mP_{\mathrm{negT}} &:= \left( \hzi^{(j)} \odot \hat{\mm} ( \hzt^{(i)} ), \; \hzt^{(i)} \right).
\end{align}
In particular, $ \mP_{\mathrm{negI}} $ contrasts the image representation $ \hzi^{(i)} $ in the positive pair with randomly sampled caption representations $ \hzt^{(j)} $ (see the \textcolor{seafoam}{green} region in Figure~\ref{fig:model_arch}), whereas $ \mP_{\mathrm{negT}} $ contrasts the text representation $ \hzt^{(i)} $ in the positive pair with randomly sampled image representations $ \hzi^{(j)}$ (see the \textcolor{orange}{orange} region in Figure~\ref{fig:model_arch}).

\paragraph{Sparsity penalty.}
We implement $\Ls$ in \eqref{eq:contrastive_objective} with a $\ell_{1}$ term for its compatibility with deep-learning training:
\begin{align} \label{eq:sparsity_term}
    \cL_{ \mathrm{sparsity} } = \norm{ \hat{\mm} (\vt) }_{1}.
\end{align}
This term ensures that the textual concepts are encoded into a minimal number of latent dimensions, promoting the disentanglement of distinct concepts across text captions. 

\paragraph{\ours training objective.}
In summary, the training objective of \ours is a weighted sum of loss terms in \eqref{eq:ours_symmetric_contrast} and \eqref{eq:sparsity_term}:
\begin{align}
    \cL &= \lambda_{\textrm{align}} \cdot \left( \cL_{ \mathrm{ctrI} } + \cL_{ \mathrm{ctrT} } \right) + \lambda_{ \textrm{sparsity}  } \cdot \cL_{ \mathrm{sparsity} },
\end{align}
where $ \lambda_{\textrm{align}} $ and $ \lambda_{ \textrm{sparsity} } $ denote the weighting coefficients.


\section{Experiments}
\subsection{Setup}

\begin{table*}
\centering
\setlength{\tabcolsep}{3pt}
\caption{
    \textbf{Results of short-caption text-image retrieval on the 5k COCO2017 validation set and the whole 30k Flickr30K dataset.} 
}
\label{tab:2}
\begin{tabular}{c|c|ccc|ccc|ccc|ccc}
\hline
& & \multicolumn{6}{c|}{COCO} & \multicolumn{6}{c}{Flickr30k} \\
& & \multicolumn{3}{c|}{Image-to-Text} & \multicolumn{3}{c|}{Text-to-Image} & \multicolumn{3}{c|}{Image-to-Text} & \multicolumn{3}{c}{Text-to-Image} \\
& & R@1 & R@5 & R@10 & R@1 & R@5 & R@10 & R@1 & R@5 & R@10 & R@1 & R@5 & R@10 \\
\hline
\multirow{3}{*}{B/16} & CLIP & 51.8 & 76.8 & 84.3 & 32.7 & 57.7 & 68.2 & 44.1 & 68.2 & 77.0 & 24.7 & 45.1 & 54.6 \\
& Direct Fine-tuning & 37.4 & 62.3 & 72.1 & 21.8 & 43.4 & 54.5 & 25.7 & 45.8 & 55.4 & 17.9 & 34.5 & 43.1 \\
& Long-CLIP \cite{zhang2024long}& {57.6} & {81.1} & {87.8} & {40.4} & {65.8} & {75.2} & {46.8} & {71.4} & {79.8} & {34.1} & {56.3} & {65.7} \\
& \ours(Ours) & \textbf{61.9} & \textbf{83.3} & \textbf{89.7} & \textbf{42.4} & \textbf{68.2} & \textbf{77.8} & \textbf{55.6} & \textbf{78.2} & \textbf{85.0} & \textbf{36.3} & \textbf{58.8} & \textbf{67.8} \\
\hline
\multirow{3}{*}{L/14} & CLIP & 56.1 & 79.5 & 86.8 & 35.4 & 60.1 & 70.2 & 48.5 & 72.6 & 80.8 & 28.0 & 49.3 & 58.7 \\
& Direct Fine-tuning & 37.9 & 63.1 & 72.2 & 23.1 & 45.1 & 55.9 & 26.0 & 46.3 & 55.6 & 17.9 & 34.9 & 43.5 \\
& Long-CLIP \cite{zhang2024long} & {62.8} & {85.1} & {91.2} & {46.3} & {70.8} & {79.8} & {53.4} & {77.5} & {85.3} & {41.2} & {64.1} & {72.6} \\
& \ours(Ours) & \textbf{66.0} & \textbf{86.2} & \textbf{92.6} & \textbf{48.5} & \textbf{73.1} & \textbf{81.7} & \textbf{63.9} & \textbf{84.4} & \textbf{90.2} & \textbf{43.8} & \textbf{66.5} & \textbf{74.8} \\
\hline
\end{tabular}
\label{tab:short}
\vspace{-0.4cm}
\end{table*}

\begin{table*}[ht]

    \begin{minipage}{0.5\textwidth}
        \setlength{\tabcolsep}{4pt}
        \caption{
            \textbf{The R@1 of long-caption text-image retrieval on 1k ShareGPT4V [2] validation set and Urban-1000 dataset.} 
            The best results are \textbf{bold}. We cite the results from Long-CLIP~\citep{zhang2024long}.
        }
        
        \begin{tabular}{c|c|cc|cc}
        \hline
        & & \multicolumn{2}{c|}{ShareGPT4V} & \multicolumn{2}{c}{Urban1k} \\
        & & I2T & T2I & I2T & T2I \\
        \hline
        \multirow{3}{*}{B/16} & CLIP  \cite{radford2021learning}& 78.2 & 79.6 & 68.1 & 53.6 \\
        & Direct Fine-tuning & 94.1 & {93.6} & -&- \\
        & Long-CLIP \cite{zhang2024long} & {94.6} & 93.3 & {78.9} & {79.5} \\
        & \ours (Ours) & \textbf{98.7} & \textbf{98.1} & \textbf{90.0} & \textbf{87.4} \\
        \hline
        \multirow{3}{*}{L/14} & CLIP \cite{radford2021learning} & 81.8 & 84.0 & 68.7 & 52.8 \\
        & Direct Fine-tuning & 95.3 & 95.4 & -&- \\
        & Long-CLIP \cite{zhang2024long}& {95.8} & {95.6} & {82.7} & {86.1} \\
        & \ours (Ours)& \textbf{97.9} & \textbf{98.5} & \textbf{93.0} & \textbf{90.1} \\
        \hline
        \end{tabular}
        
        \label{tab:long}
    \end{minipage}
    \hfill
     \begin{minipage}{0.45\textwidth}
        \centering
        \caption{
            \textbf{
                Zero-shot classification performance on ViT-L/14 models.} 
                When the class name is very short, i.e., a single word like ImageNet, CLIP model perform better. 
                When the class name is a combination of several words, our method achieves better results, e.g., the road sign in GTSRB.
        }
        \begin{tabular}{c|ccc}
        \hline
        Dataset & CLIP  & LongCLIP  & \ours \\ \hline
        Country211 & \textbf{31.8} &28.1 & 26.9\\
            Fer2013 & 49.0 & 57.8 & \textbf{58.6} \\ 
            Fgvc-aircraft & \textbf{31.7} & 30.6 & 30.4\\
            GTSRB & 50.2 & 48.9 & \textbf{52.4}\\
            ImageNet & \textbf{75.3} &72.9 & 72.5  \\
            ImageNet-V2 & \textbf{69.7} &66.9 & 66.6  \\
            VOC2007 & 78.3 & 77.5 & \textbf{78.6}\\
            VOC2007-Multi & 79.0 & 82.1 & \textbf{83.7}\\
            SUN397 & 67.5 & \textbf{72.5}&72.1\\ \hline
        \end{tabular}
        \label{tab:zeroshot}
    \end{minipage}
\end{table*}

\paragraph{Implementation details.}
Following Long-CLIP~\citep{zhang2024long}, we finetune the CLIP model \cite{radford2021learning} on ShareGPT4V~\citep{chen2023sharegpt4v}, which contains around 1 million image-text pairs. 
We employ the position encoding in long-CLIP to handle 248 tokens (c.f., the 77-token limit in the original CLIP).
Compared to the baseline CLIP model, we introduce a mask network $\hat{\mm}$. 
The masking network is designed as a single transformer block, which takes the text sequence embedding $\hzt$ from the text encoder. Then we add an attention-pooling layer to down-sample it to the same size as the CLIP representation, e.g., 768 in ViT-L/14. We tested including more transformer blocks in the mask network but did not observe significant improvements. 
Therefore, we stick to one block for faster training and inference.
Unlike Long-CLIP \cite{zhang2024long} which processes all the captions for each image at each gradient step, we only sample one caption for each image, reducing our overall training time by half.
Specifically, on 8 H100 GPUs, training a Vit-B/16 model for one epoch takes about 4 minutes with our model, whereas it takes around 7 minutes for Long-CLIP. After the pooling layer, we apply sigmoid to restrict the output to the range $(0,1)$ and employ straight-through estimation~\citep{bengio2013estimating} to binarize the outputs. 
The training batch size is 1024 and the learning rate is $10^{-6}$ for the CLIP component and $10^{-3}$ for the mask network.

\paragraph{Evaluation.}
We evaluate the following datasets: 
\begin{itemize}
    \item Long text-to-image retrieval datasets: ShareGPT4V validation split~\citep{chen2023sharegpt4v} and Urban1k~\cite{zhang2024long}. The captions for each image are long and describe details about the image. Both datasets contain 1000 text-to-image pairs.
    \item Short text-to-image retrieval datasets: COCO2017 validation split~\cite{lin2014microsoft} and Flick30K~\cite{young2014image}. Following Long-CLIP~\citep{zhang2024long}, we use 30K Flickr training dataset. 
    \item Zero-shot image classification datasets. We use benchmark datasets: Country211, Fer2013, Fgvc-aircraft, GTSRB, ImageNet, ImagetNet-V2, VOC2007, VOC2007-Multi, and SUN397. \footnote{\url{https://github.com/LAION-AI/CLIP_benchmark}}
\end{itemize}

\paragraph{Baselines.}
In this paper, we benchmark our approach against CLIP \cite{radford2021learning} and the recent state-of-the-art model long-CLIP \cite{zhang2024long}.

\begin{figure*}
\hrule
\vspace{0.1cm}
    \centering
    \begin{minipage}{0.6\textwidth}
    \small
    \emph{This playful food sculpture transforms cucumbers into a fearsome T-Rex dinosaur. 
    The \textcolor{darkgreen}{\textbf{\emph{cucumbers form the main body}}}, with whole cucumbers creating the legs and tail, 
    while sliced cucumbers make up the creature's midsection. More cucumbers are cleverly 
    cut to shape the dinosaur's head, and additional cucumbers are arranged to suggest 
    muscular limbs.
    From its mouth erupts a dramatic spray of {carrots}, with finely julienned \textcolor{orange}{\textbf{\emph{carrots
    creating the effect of fire}}}. These bright orange carrots provide a stunning contrast 
    against the green vegetables. Shredded carrots cascade downward like flames, while 
    more carrots are delicately cut to create a flame-like texture. The carrots' vibrant 
    color makes the dinosaur appear truly animated.
    Fresh celery leaves crown the creation, with celery fronds adding a decorative touch 
    around the body. More \textcolor{seafoam}{\textbf{\emph{celery leaves create a natural backdrop}}}, while additional celery 
    pieces add texture throughout. The celery's feathered appearance provides an artistic 
    flourish to the overall design.} 
    \end{minipage}%
    \hfill
     \setlength{\tabcolsep}{1pt}
    \begin{minipage}{0.35\textwidth}
    \begin{tabular}{cc}
    \rotatebox{90}{\scriptsize ~~~~~ \textbf{CLIP}} &
    \includegraphics[scale=0.065]{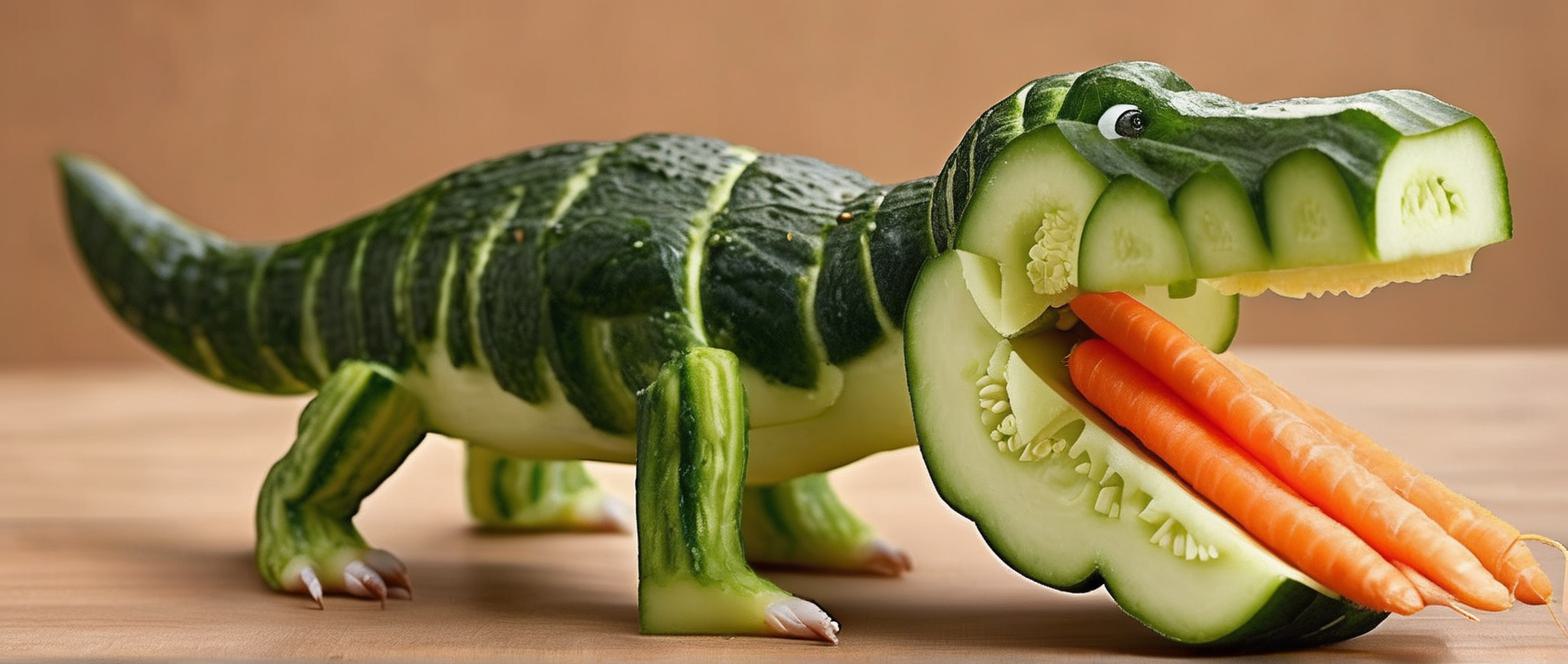} \\[0ex]
    \rotatebox{90}{\scriptsize ~~~~~ \textbf{Long-CLIP}}  &
    \includegraphics[scale=0.065]{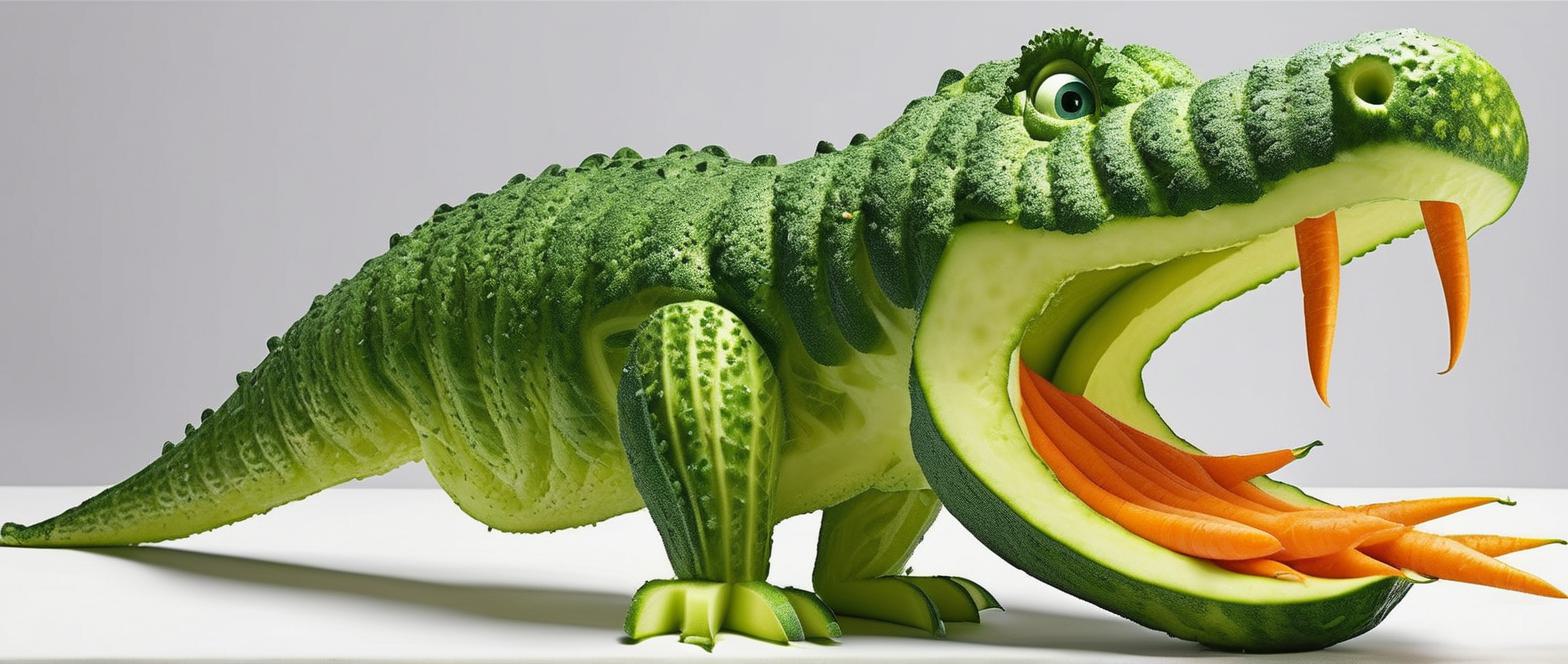} \\[0ex]
    \rotatebox{90}{\scriptsize ~~~~~ \ours} &
    \includegraphics[scale=0.065]{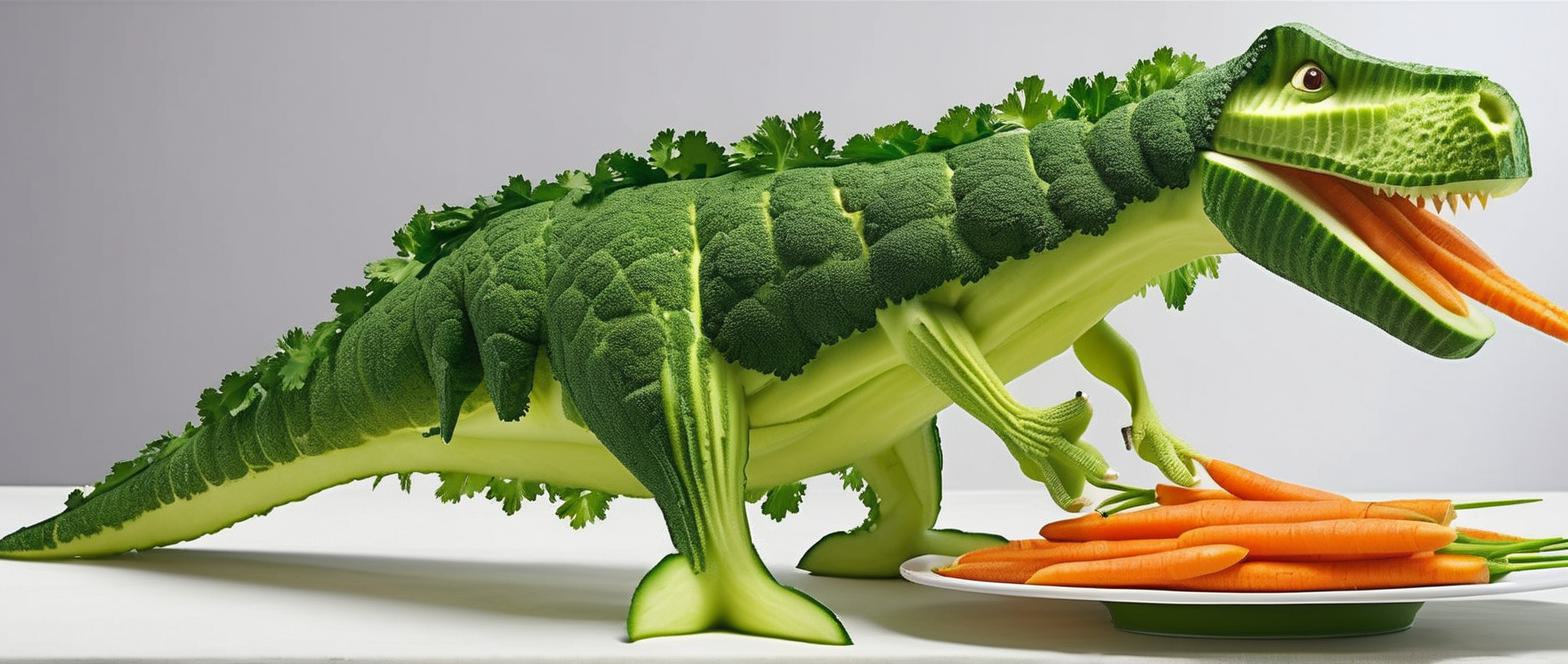}
    \end{tabular}
    \end{minipage}
    \hrule
    \caption{
        \textbf{Example of Long-text-to-image generation.} 
        We replace the CLIP text encoder in SDXL with different finetuned CLIP models. Given a very long text, CLIP \citep{radford2021learning} truncates the input to 77 tokens, resulting in information loss in the image. Our model learns to generate details such as celery leaves on the back of the dinosaur while other models fail. 
    }
    \label{fig:example_t2i}
    \vspace{-0.7cm}
\end{figure*}

\begin{figure*}[ht]
    \centering
     \begin{subfigure}[b]{0.24\textwidth}
        \centering
        \includegraphics[width=4cm, height=3cm]{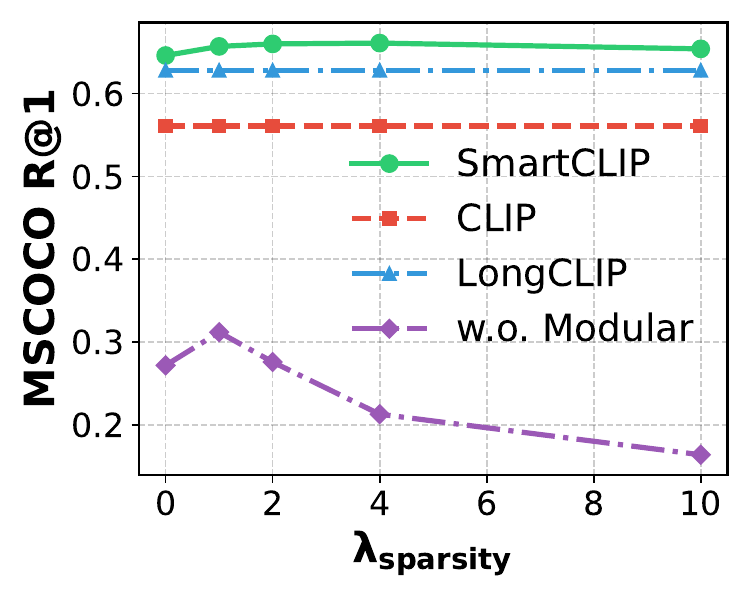}
        \label{fig:first_sparsity}
    \end{subfigure}
    \hfill
    \begin{subfigure}[b]{0.24\textwidth}
        \centering
        \includegraphics[width=4cm, height=3cm]{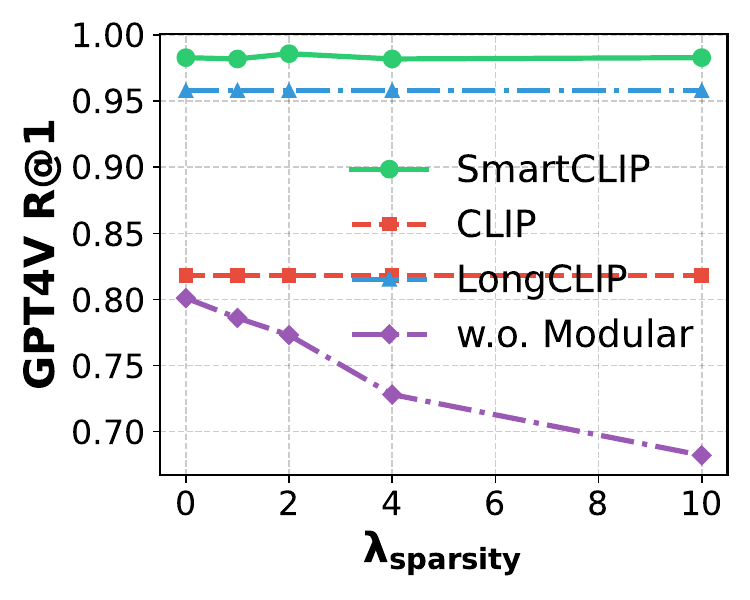}
        \label{fig:second_sparsity}
    \end{subfigure}
    \hfill
    \begin{subfigure}[b]{0.24\textwidth}
        \centering
        \includegraphics[width=4cm, height=3cm]{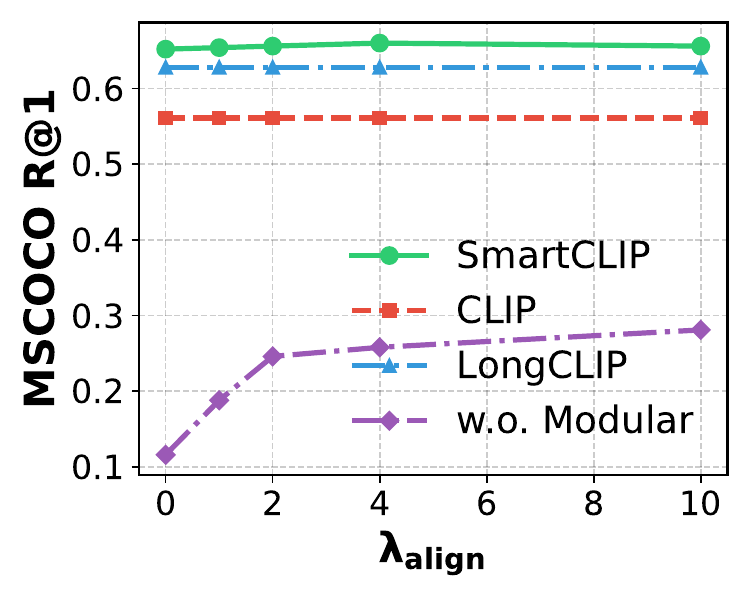}
        \label{fig:first_align}
    \end{subfigure}
    \hfill
    \begin{subfigure}[b]{0.24\textwidth}
        \centering
        \includegraphics[width=4cm, height=3cm]{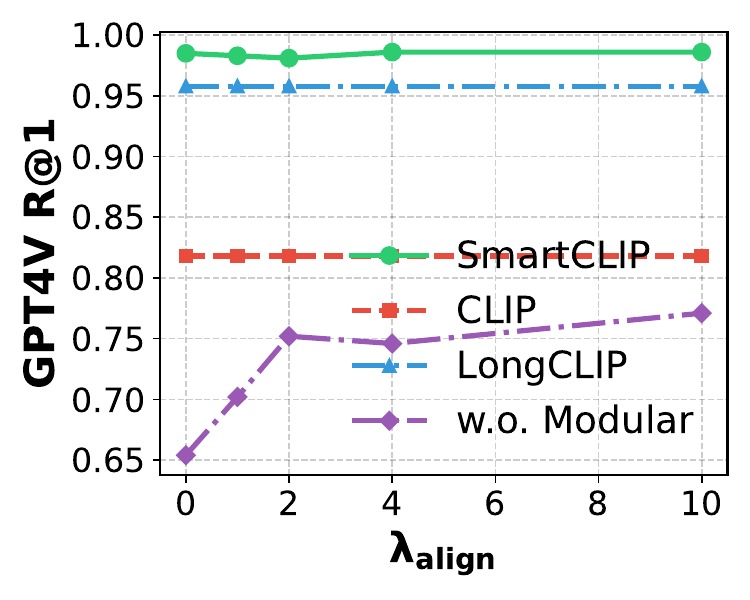}
        \label{fig:second_align}
    \end{subfigure}
    \vspace{-0.3cm}
    \caption{
        \textbf{Ablation Studies on two proposed modules: selective alignment and sparsity.} 
        The baseline \emph{w.o. Modular} means that we replace our modular alignment module with standard contrastive learning alignment.}
    \label{fig:ablation}
    \vspace{-0.5cm}
\end{figure*}

\begin{figure}
    \centering
   \setlength{\tabcolsep}{1pt}
   \begin{tabular}{ccccc}
   &Input & CLIP & LongCLIP & \ours \\
      \rotatebox{90}{~~~~~ \textbf{Deer}} & \includegraphics[scale=0.23]{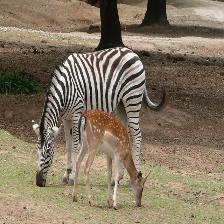}  &
      \includegraphics[scale=0.23]{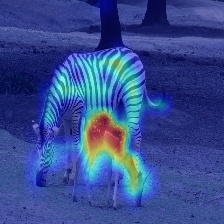} &
       \includegraphics[scale=0.23]{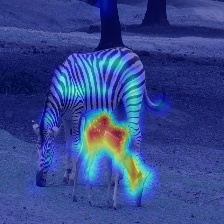} &
        \includegraphics[scale=0.23]{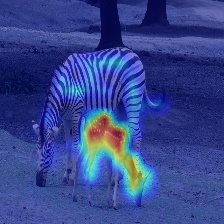} 
      \\
        \rotatebox{90}{~ \textbf{skateboard}} & \includegraphics[scale=0.23]{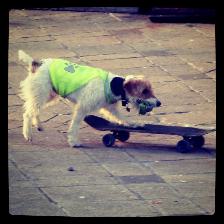}  &
      \includegraphics[scale=0.23]{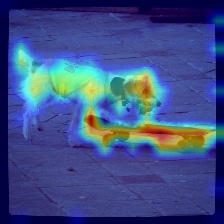} &
       \includegraphics[scale=0.23]{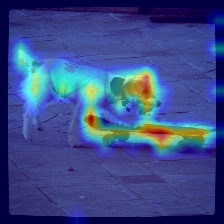} &
        \includegraphics[scale=0.23]{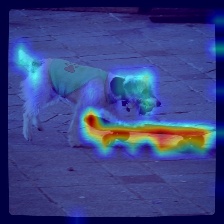} 
   \end{tabular}
    \caption{
        \textbf{Visualization of learned representations.} 
        Given an image, we generate two captions (e.g., \emph{a zebra} and \emph{a zebra and a deer}) and compute cosine similarities with the same image embedding to perform binary classification for visualization using ScoreCAM \cite{wang2020score}. Compared to baseline methods, our CLIP representations are more atomic and capture differences more effectively.
    }
    \label{fig:visualization}
    \vspace{-0.3cm}
\end{figure}



\subsection{Comparison with CLIP Models}
We present our experimental results across three key tasks: long text-to-image retrieval, short text-to-image retrieval, and zero-shot classification.

\paragraph{Long text-to-image retrieval.}
Table~\ref{tab:long} showcases our method's performance on long text-to-image retrieval tasks. \ours achieves substantial improvements over baseline models, particularly the SOTA Long-CLIP, which is designed to handle long text sequences. For example, on the Urban1k dataset, \ours elevates the performance from 78.9\% to 90.0\%, marking an impressive 14\% boost.

\paragraph{Short text-to-image retrieval.}
Similarly, as illustrated in Table~\ref{tab:short}, \ours significantly outperforms all baseline models across various metrics and datasets in short text-to-image retrieval tasks. The encouraging performance gains show that \ours can capture detailed information within images while simultaneously emphasizing the main semantic content.

\paragraph{Zero-shot classification.}
To comprehensively evaluate our model's capabilities, we conduct zero-shot classification benchmarks in Table~\ref{tab:zeroshot}. Both the standard CLIP model and \ours demonstrate superior performance on different datasets. Notably, \ours exhibits a slight performance decline on ImageNet, which is expected since our model is fine-tuned on the ShareGPT4V dataset featuring long text captions, whereas ImageNet primarily consists of short, often single-word class names. However, \ours excels on datasets with multi-word class names, such as the GTSRB dataset, where it achieves the best performance in accurately classifying road sign descriptions.


\subsection{Ablation Studies}
We analyze the three key components in our model: the modular alignment module, the sparsity loss and the impact of caption diversity in the data.

\paragraph{Modular alignment.}  
After introducing the mask network, we replace the standard contrastive learning with our selective alignment module. 
As shown in Figure~\ref{fig:ablation}, this change significantly improves performance. When using standard contrastive learning instead of the modular contrastive module (indicated by the purple lines), performance drops sharply. This happens because the mask information allows the network to easily separate positive pairs from negative ones, making the negative samples less informative. As a result, standard contrastive learning no longer effectively helps the model learn meaningful information.

\paragraph{Alignment coefficient $\lambda_{\mathrm{align}}$.}  
We test the impact of the alignment coefficient $\lambda_{\mathrm{align}}$. 
The results in the right panels of Figure~\ref{fig:ablation} show that our method performs consistently well across a wide range of $\lambda_{\mathrm{align}}$ values, from 0.1 to 20. This indicates that our approach is robust and does not require precise tuning of $\lambda_{\mathrm{align}}$ to achieve good performance.

\paragraph{Sparsity coefficient $\lambda_{\mathrm{sparsity}}$.}  
We also examine the sparsity coefficient $\lambda_{\mathrm{sparsity}}$.
The left panels of Figure~\ref{fig:ablation} demonstrate that adding sparsity to the mask network improves performance. This supports our idea that promoting sparsity helps the model focus on the most relevant concepts, enhancing its ability to capture detailed information without being distracted by irrelevant details.

\paragraph{Caption diversity.} We evaluate our model's performance under varying caption diversity conditions using the COCO dataset \cite{lin2014microsoft}. As shown in Table~\ref{tab:cap_div}, increasing the number of captions per image enhances performance on the Flickr30K dataset, though at the expense of degraded performance on long-text-to-image retrieval tasks. Further improvements are achieved when we combine our training dataset with COCO. These results highlight both the importance of caption diversity and our method's capability to effectively handle complex text-image datasets.

\subsection{Additional Results} \label{subsec:addition_results}

\paragraph{Visualization.}
While our quantitative results demonstrate superior performance across various tasks, we also explore the qualitative aspects of our model by visualizing the learned representations. 
The visualization results are shown in Figure~\ref{fig:visualization}.
We employ the ScoreCAM method \cite{wang2020score} for this purpose. For each image, we generate two distinct captions, such as \emph{``a zebra''} and \emph{``a zebra and a deer''}. We then compute the cosine similarity between the image embedding and each of the two text embeddings. These similarity scores serve as logits for a classification task, which are then input into the ScoreCAM algorithm. 
\ours successfully learns modular representations, accurately capturing the relevant differences between captions. 
\begin{table}[]
    \centering
    \small
     \vspace{-0.25cm}
     \renewcommand\arraystretch{1.0}
\renewcommand\tabcolsep{3pt}
    \begin{tabular}{ccccc}
    \hline
    NumCapPerImg-COCO & T2I & I2T & LongT2I & LongI2T \\ \hline
         1&  53.6 & 39.3  & 85.3 & 89.3\\ 
         3 & 53.6 & 40.9 & 85.3&88.6\\ 
         5 & 56.4 & 41.2  &85.2 & 86.5 \\ \hline
         Ours-ShareGPT & 55.6 & 36.3 & 98.7 & 98.1 \\ \hline
         +COCO & 57.0 &38.4 & 97.8 & 98.5\\ \hline
    \end{tabular}
    \vspace{-0.3cm}
    \caption{\small Retrieval results on Flickr30K \cite{young2014image} and ShareGPT-val \cite{chen2023sharegpt4v} with models trained with different caption counts per image (top) and mixing long and short text-image datasets (bottom).}
    \label{tab:cap_div}
    \vspace{-0.2cm}
\end{table}

\paragraph{Plug and play for text-to-image generation.}
One main advantage of \ours over other CLIP models trained from scratch is the low computational cost of finetuning. Additionally, our fine-tuned text encoder can replace the CLIP text encoders in large-scale models in a plug-and-play manner. 
Specifically, we substitute the text encoder in the SDXL~\cite{podell2023sdxl} model with both Long-CLIP and \ours.
As illustrated in Figure~\ref{fig:example_t2i}, our text encoder demonstrates a superior understanding of the long text, generating detailed elements such as celery leaves in the background. Furthermore, Table~\ref{tab:t2i} presents quantitative results on image generation from long captions in the ShareGPT4V validation split. Our method consistently achieves better performance across various metrics, showcasing its effectiveness in handling complex, long text inputs.

\begin{table}
 \setlength{\tabcolsep}{4pt}
    \centering
    \caption{
        \textbf{Long text to image generation performance.} 
        We use the long captions (usually around 200 tokens, beyond the hard constraint 77 of the original CLIP model) to generate images with SDXL model \cite{podell2023sdxl}, then we compare the generated images against the real images in ShareGPT4V validation split.
    }
    \begin{tabular}{cccccc}
    \hline
        Method & KID $\downarrow$ & Pr $\uparrow$ & Re $\uparrow$ & F1 $\uparrow$ & DINO-L $\uparrow$ \\ \hline
        LongCLIP &  1.05 & 0.238 & 0.768 & 0.363 & 0.401\\
        \ours & \textbf{1.02} & \textbf{0.258}& \textbf{0.791} & \textbf{0.389} &  \textbf{0.414}\\ \hline
    \end{tabular}
    \label{tab:t2i}
    \vspace{-0.5cm}
\end{table}

\section{Conclusion and Limitation}
In this work, we address the information misalignment and  representation entanglement issues in existing vision-language models (e.g., CLIP). 
We establish theoretical conditions for effectively connecting text representations to atomic-level visual features and propose \ours, a principled, refined vision-language model. 
Our experimental results validate both our theoretical results and the practical effectiveness of \ours in advancing multimodal learning.
\paragraph{Limitation.}
As discussed earlier, Condition~\ref{cond:identification_conditions}-\ref{asmp:full_support} could be violated for datasets in which a subset images are paired with a limited number of captions compared to others.
In Section~\ref{sec:theory}, we discuss practical strategies to mitigate such issues.
Devising alternative theoretical conditions may provide additional insights into fully utilizing all pairing information, which we leave as future work.

\section*{Acknowledgement}
We would like to acknowledge the support from NSF Award No. 2229881, AI Institute for Societal Decision Making (AI-SDM), the National Institutes of Health (NIH) under Contract R01HL159805, and grants from Quris AI, Florin Court Capital, and MBZUAI-WIS Joint Program.
The work of L. Kong is supported in part by NSF DMS-2134080 through an award to Y. Chi.

{
    \small
    \bibliographystyle{ieeenat_fullname}
    \bibliography{references}

@misc{hyvarinen2016unsupervised,
      title={Unsupervised Feature Extraction by Time-Contrastive Learning and Nonlinear ICA}, 
      author={Aapo Hyvarinen and Hiroshi Morioka},
      year={2016},
      eprint={1605.06336},
      archivePrefix={arXiv},
      primaryClass={stat.ML}
}

@inproceedings{hyvarinen2019nonlinear,
  title={Nonlinear ICA using auxiliary variables and generalized contrastive learning},
  author={Hyvarinen, Aapo and Sasaki, Hiroaki and Turner, Richard},
  booktitle={The 22nd International Conference on Artificial Intelligence and Statistics},
  pages={859--868},
  year={2019},
  organization={PMLR}
}

@inproceedings{khemakhem2020variational,
  title={Variational autoencoders and nonlinear ica: A unifying framework},
  author={Khemakhem, Ilyes and Kingma, Diederik and Monti, Ricardo and Hyvarinen, Aapo},
  booktitle={International Conference on Artificial Intelligence and Statistics},
  pages={2207--2217},
  year={2020},
  organization={PMLR}
}

@article{von2021self,
  title={Self-Supervised Learning with Data Augmentations Provably Isolates Content from Style},
  author={von K{\"u}gelgen, Julius and Sharma, Yash and Gresele, Luigi and Brendel, Wieland and Sch{\"o}lkopf, Bernhard and Besserve, Michel and Locatello, Francesco},
  journal={arXiv preprint arXiv:2106.04619},
  year={2021}
}

@InProceedings{kong2022partial,
  title = 	 {Partial disentanglement for domain adaptation},
  author =       {Kong, Lingjing and Xie, Shaoan and Yao, Weiran and Zheng, Yujia and Chen, Guangyi and Stojanov, Petar and Akinwande, Victor and Zhang, Kun},
  booktitle = 	 {Proceedings of the 39th International Conference on Machine Learning},
  pages = 	 {11455--11472},
  year = 	 {2022},
  editor = 	 {Chaudhuri, Kamalika and Jegelka, Stefanie and Song, Le and Szepesvari, Csaba and Niu, Gang and Sabato, Sivan},
  volume = 	 {162},
  series = 	 {Proceedings of Machine Learning Research},
  month = 	 {17--23 Jul},
  publisher =    {PMLR},
  url = 	 {https://proceedings.mlr.press/v162/kong22a.html},
}

@misc{sturma2023unpaired,
      title={Unpaired Multi-Domain Causal Representation Learning}, 
      author={Nils Sturma and Chandler Squires and Mathias Drton and Caroline Uhler},
      year={2023},
      eprint={2302.00993},
      archivePrefix={arXiv},
      primaryClass={stat.ML}
}

@misc{lachapelle2023additive,
      title={Additive Decoders for Latent Variables Identification and Cartesian-Product Extrapolation}, 
      author={Sébastien Lachapelle and Divyat Mahajan and Ioannis Mitliagkas and Simon Lacoste-Julien},
      year={2023},
      eprint={2307.02598},
      archivePrefix={arXiv},
      primaryClass={cs.LG}
}

@inproceedings{rombach2022high,
  title={High-resolution image synthesis with latent diffusion models},
  author={Rombach, Robin and Blattmann, Andreas and Lorenz, Dominik and Esser, Patrick and Ommer, Bj{\"o}rn},
  booktitle={Proceedings of the IEEE/CVF conference on computer vision and pattern recognition},
  pages={10684--10695},
  year={2022}
}

@article{von2024nonparametric,
  title={Nonparametric identifiability of causal representations from unknown interventions},
  author={von K{\"u}gelgen, Julius and Besserve, Michel and Wendong, Liang and Gresele, Luigi and Keki{\'c}, Armin and Bareinboim, Elias and Blei, David and Sch{\"o}lkopf, Bernhard},
  journal={Advances in Neural Information Processing Systems},
  volume={36},
  year={2024}
}

@article{zhang2024identifiability,
  title={Identifiability guarantees for causal disentanglement from soft interventions},
  author={Zhang, Jiaqi and Greenewald, Kristjan and Squires, Chandler and Srivastava, Akash and Shanmugam, Karthikeyan and Uhler, Caroline},
  journal={Advances in Neural Information Processing Systems},
  volume={36},
  year={2024}
}

@article{buchholz2024learning,
  title={Learning linear causal representations from interventions under general nonlinear mixing},
  author={Buchholz, Simon and Rajendran, Goutham and Rosenfeld, Elan and Aragam, Bryon and Sch{\"o}lkopf, Bernhard and Ravikumar, Pradeep},
  journal={Advances in Neural Information Processing Systems},
  volume={36},
  year={2024}
}

@article{ramesh2022hierarchical,
  title={Hierarchical text-conditional image generation with clip latents},
  author={Ramesh, Aditya and Dhariwal, Prafulla and Nichol, Alex and Chu, Casey and Chen, Mark},
  journal={arXiv preprint arXiv:2204.06125},
  volume={1},
  number={2},
  pages={3},
  year={2022}
}

@article{zhang2024causal,
  title={Causal representation learning from multiple distributions: A general setting},
  author={Zhang, Kun and Xie, Shaoan and Ng, Ignavier and Zheng, Yujia},
  journal={arXiv preprint arXiv:2402.05052},
  year={2024}
}

@inproceedings{radford2021learning,
  title={Learning transferable visual models from natural language supervision},
  author={Radford, Alec and Kim, Jong Wook and Hallacy, Chris and Ramesh, Aditya and Goh, Gabriel and Agarwal, Sandhini and Sastry, Girish and Askell, Amanda and Mishkin, Pamela and Clark, Jack and others},
  booktitle={International conference on machine learning},
  pages={8748--8763},
  year={2021},
  organization={PMLR}
}

@article{fan2024improving,
  title={Improving clip training with language rewrites},
  author={Fan, Lijie and Krishnan, Dilip and Isola, Phillip and Katabi, Dina and Tian, Yonglong},
  journal={Advances in Neural Information Processing Systems},
  volume={36},
  year={2024}
}

@inproceedings{mu2022slip,
  title={Slip: Self-supervision meets language-image pre-training},
  author={Mu, Norman and Kirillov, Alexander and Wagner, David and Xie, Saining},
  booktitle={European conference on computer vision},
  pages={529--544},
  year={2022},
  organization={Springer}
}

@article{zheng2024dreamlip,
  title={DreamLIP: Language-Image Pre-training with Long Captions},
  author={Zheng, Kecheng and Zhang, Yifei and Wu, Wei and Lu, Fan and Ma, Shuailei and Jin, Xin and Chen, Wei and Shen, Yujun},
  journal={arXiv preprint arXiv:2403.17007},
  year={2024}
}

@article{zhang2024long,
  title={Long-clip: Unlocking the long-text capability of clip},
  author={Zhang, Beichen and Zhang, Pan and Dong, Xiaoyi and Zang, Yuhang and Wang, Jiaqi},
  journal={arXiv preprint arXiv:2403.15378},
  year={2024}
}

@String(AAAI = {AAAI})

@inproceedings{jia2021scaling,
  title={Scaling up visual and vision-language representation learning with noisy text supervision},
  author={Jia, Chao and Yang, Yinfei and Xia, Ye and Chen, Yi-Ting and Parekh, Zarana and Pham, Hieu and Le, Quoc and Sung, Yun-Hsuan and Li, Zhen and Duerig, Tom},
  booktitle={International conference on machine learning},
  pages={4904--4916},
  year={2021},
  organization={PMLR}
}

@inproceedings{li2022blip,
  title={Blip: Bootstrapping language-image pre-training for unified vision-language understanding and generation},
  author={Li, Junnan and Li, Dongxu and Xiong, Caiming and Hoi, Steven},
  booktitle={International conference on machine learning},
  pages={12888--12900},
  year={2022},
  organization={PMLR}
}

@article{lai2024veclip,
  title={VeCLIP: Improving CLIP Training via Visual-enriched Captions},
  author={Lai, Zhengfeng and Zhang, Haotian and Wu, Wentao and Bai, Haoping and Timofeev, Aleksei and Du, Xianzhi and Gan, Zhe and Shan, Jiulong and Chuah, Chen-Nee and Yang, Yinfei and others},
  journal={ECCV. IEEE},
  pages={13},
  year={2024},
  publisher={Springer}
}

@inproceedings{li2023blip,
  title={Blip-2: Bootstrapping language-image pre-training with frozen image encoders and large language models},
  author={Li, Junnan and Li, Dongxu and Savarese, Silvio and Hoi, Steven},
  booktitle={International conference on machine learning},
  pages={19730--19742},
  year={2023},
  organization={PMLR}
}

@article{li2021supervision,
  title={Supervision exists everywhere: A data efficient contrastive language-image pre-training paradigm},
  author={Li, Yangguang and Liang, Feng and Zhao, Lichen and Cui, Yufeng and Ouyang, Wanli and Shao, Jing and Yu, Fengwei and Yan, Junjie},
  journal={arXiv preprint arXiv:2110.05208},
  year={2021}
}

@article{yu2022coca,
  title={Coca: Contrastive captioners are image-text foundation models},
  author={Yu, Jiahui and Wang, Zirui and Vasudevan, Vijay and Yeung, Legg and Seyedhosseini, Mojtaba and Wu, Yonghui},
  journal={arXiv preprint arXiv:2205.01917},
  year={2022}
}

@inproceedings{zhai2022lit,
  title={Lit: Zero-shot transfer with locked-image text tuning},
  author={Zhai, Xiaohua and Wang, Xiao and Mustafa, Basil and Steiner, Andreas and Keysers, Daniel and Kolesnikov, Alexander and Beyer, Lucas},
  booktitle={Proceedings of the IEEE/CVF conference on computer vision and pattern recognition},
  pages={18123--18133},
  year={2022}
}

@inproceedings{zhai2023sigmoid,
  title={Sigmoid loss for language image pre-training},
  author={Zhai, Xiaohua and Mustafa, Basil and Kolesnikov, Alexander and Beyer, Lucas},
  booktitle={Proceedings of the IEEE/CVF International Conference on Computer Vision},
  pages={11975--11986},
  year={2023}
}

@article{li2024if,
  title={What If We Recaption Billions of Web Images with LLaMA-3?},
  author={Li, Xianhang and Tu, Haoqin and Hui, Mude and Wang, Zeyu and Zhao, Bingchen and Xiao, Junfei and Ren, Sucheng and Mei, Jieru and Liu, Qing and Zheng, Huangjie and others},
  journal={arXiv preprint arXiv:2406.08478},
  year={2024}
}

@article{meta2024introducing,
  title={Introducing meta llama 3: The most capable openly available llm to date},
  author={Meta, AI},
  journal={Meta AI},
  year={2024}
}

@article{lai2024revisit,
  title={Revisit Large-Scale Image-Caption Data in Pre-training Multimodal Foundation Models},
  author={Lai, Zhengfeng and Saveris, Vasileios and Chen, Chen and Chen, Hong-You and Zhang, Haotian and Zhang, Bowen and Tebar, Juan Lao and Hu, Wenze and Gan, Zhe and Grasch, Peter and others},
  journal={arXiv preprint arXiv:2410.02740},
  year={2024}
}

@article{wu2024lotlip,
  title={LoTLIP: Improving Language-Image Pre-training for Long Text Understanding},
  author={Wu, Wei and Zheng, Kecheng and Ma, Shuailei and Lu, Fan and Guo, Yuxin and Zhang, Yifei and Chen, Wei and Guo, Qingpei and Shen, Yujun and Zha, Zheng-Jun},
  journal={arXiv preprint arXiv:2410.05249},
  year={2024}
}

@article{najdenkoska2024tulip,
  title={TULIP: Token-length Upgraded CLIP},
  author={Najdenkoska, Ivona and Derakhshani, Mohammad Mahdi and Asano, Yuki M and van Noord, Nanne and Worring, Marcel and Snoek, Cees GM},
  journal={arXiv preprint arXiv:2410.10034},
  year={2024}
}

@inproceedings{baldrati2023zero,
  title={Zero-shot composed image retrieval with textual inversion},
  author={Baldrati, Alberto and Agnolucci, Lorenzo and Bertini, Marco and Del Bimbo, Alberto},
  booktitle={Proceedings of the IEEE/CVF International Conference on Computer Vision},
  pages={15338--15347},
  year={2023}
}

@inproceedings{che2023enhancing,
  title={Enhancing Multimodal Understanding with CLIP-Based Image-to-Text Transformation},
  author={Che, Chang and Lin, Qunwei and Zhao, Xinyu and Huang, Jiaxin and Yu, Liqiang},
  booktitle={Proceedings of the 2023 6th International Conference on Big Data Technologies},
  pages={414--418},
  year={2023}
}

@inproceedings{zhang2022pointclip,
  title={Pointclip: Point cloud understanding by clip},
  author={Zhang, Renrui and Guo, Ziyu and Zhang, Wei and Li, Kunchang and Miao, Xupeng and Cui, Bin and Qiao, Yu and Gao, Peng and Li, Hongsheng},
  booktitle={Proceedings of the IEEE/CVF conference on computer vision and pattern recognition},
  pages={8552--8562},
  year={2022}
}

@inproceedings{wang2023exploring,
  title={Exploring clip for assessing the look and feel of images},
  author={Wang, Jianyi and Chan, Kelvin CK and Loy, Chen Change},
  booktitle={Proceedings of the AAAI Conference on Artificial Intelligence},
  volume={37},
  number={2},
  pages={2555--2563},
  year={2023}
}

@inproceedings{tang2021clip4caption,
  title={Clip4caption: Clip for video caption},
  author={Tang, Mingkang and Wang, Zhanyu and Liu, Zhenhua and Rao, Fengyun and Li, Dian and Li, Xiu},
  booktitle={Proceedings of the 29th ACM International Conference on Multimedia},
  pages={4858--4862},
  year={2021}
}

@inproceedings{ju2022prompting,
  title={Prompting visual-language models for efficient video understanding},
  author={Ju, Chen and Han, Tengda and Zheng, Kunhao and Zhang, Ya and Xie, Weidi},
  booktitle={European Conference on Computer Vision},
  pages={105--124},
  year={2022},
  organization={Springer}
}

@inproceedings{rasheed2023fine,
  title={Fine-tuned clip models are efficient video learners},
  author={Rasheed, Hanoona and Khattak, Muhammad Uzair and Maaz, Muhammad and Khan, Salman and Khan, Fahad Shahbaz},
  booktitle={Proceedings of the IEEE/CVF Conference on Computer Vision and Pattern Recognition},
  pages={6545--6554},
  year={2023}
}

@article{liu2024visual,
  title={Visual instruction tuning},
  author={Liu, Haotian and Li, Chunyuan and Wu, Qingyang and Lee, Yong Jae},
  journal={Advances in neural information processing systems},
  volume={36},
  year={2024}
}

@inproceedings{lin2014microsoft,
  title={Microsoft coco: Common objects in context},
  author={Lin, Tsung-Yi and Maire, Michael and Belongie, Serge and Hays, James and Perona, Pietro and Ramanan, Deva and Doll{\'a}r, Piotr and Zitnick, C Lawrence},
  booktitle={Computer Vision--ECCV 2014: 13th European Conference, Zurich, Switzerland, September 6-12, 2014, Proceedings, Part V 13},
  pages={740--755},
  year={2014},
  organization={Springer}
}

@article{chen2023sharegpt4v,
  title={Sharegpt4v: Improving large multi-modal models with better captions},
  author={Chen, Lin and Li, Jinsong and Dong, Xiaoyi and Zhang, Pan and He, Conghui and Wang, Jiaqi and Zhao, Feng and Lin, Dahua},
  journal={arXiv preprint arXiv:2311.12793},
  year={2023}
}

@inproceedings{wang2020score,
  title={Score-CAM: Score-weighted visual explanations for convolutional neural networks},
  author={Wang, Haofan and Wang, Zifan and Du, Mengnan and Yang, Fan and Zhang, Zijian and Ding, Sirui and Mardziel, Piotr and Hu, Xia},
  booktitle={Proceedings of the IEEE/CVF conference on computer vision and pattern recognition workshops},
  pages={24--25},
  year={2020}
}

@inproceedings{yaomulti,
  title={Multi-View Causal Representation Learning with Partial Observability},
  author={Yao, Dingling and Xu, Danru and Lachapelle, Sebastien and Magliacane, Sara and Taslakian, Perouz and Martius, Georg and von K{\"u}gelgen, Julius and Locatello, Francesco},
  booktitle={The Twelfth International Conference on Learning Representations},
  year={2024},
}

@inproceedings{wang2020understanding,
  title={Understanding contrastive representation learning through alignment and uniformity on the hypersphere},
  author={Wang, Tongzhou and Isola, Phillip},
  booktitle={International conference on machine learning},
  pages={9929--9939},
  year={2020},
  organization={PMLR}
}

@article{oord2018representation,
  title={Representation learning with contrastive predictive coding},
  author={Oord, Aaron van den and Li, Yazhe and Vinyals, Oriol},
  journal={arXiv preprint arXiv:1807.03748},
  year={2018}
}

@inproceedings{chen2020simple,
  title={A simple framework for contrastive learning of visual representations},
  author={Chen, Ting and Kornblith, Simon and Norouzi, Mohammad and Hinton, Geoffrey},
  booktitle={International conference on machine learning},
  pages={1597--1607},
  year={2020},
  organization={PMLR}
}

@inproceedings{daunhaweridentifiability,
  title={Identifiability Results for Multimodal Contrastive Learning},
  author={Daunhawer, Imant and Bizeul, Alice and Palumbo, Emanuele and Marx, Alexander and Vogt, Julia E},
  booktitle={The Eleventh International Conference on Learning Representations},
  year={2023}
}

@inproceedings{morioka2023connectivity,
  title={Connectivity-contrastive learning: Combining causal discovery and representation learning for multimodal data},
  author={Morioka, Hiroshi and Hyvarinen, Aapo},
  booktitle={International conference on artificial intelligence and statistics},
  pages={3399--3426},
  year={2023},
  organization={PMLR}
}

@inproceedings{morioka2024causal,
  title={Causal Representation Learning Made Identifiable by Grouping of Observational Variables},
  author={Morioka, Hiroshi and Hyvarinen, Aapo},
  booktitle={Forty-first International Conference on Machine Learning},
  year={2024},
}

@inproceedings{gresele2020incomplete,
  title={The incomplete rosetta stone problem: Identifiability results for multi-view nonlinear ica},
  author={Gresele, Luigi and Rubenstein, Paul K and Mehrjou, Arash and Locatello, Francesco and Sch{\"o}lkopf, Bernhard},
  booktitle={Uncertainty in Artificial Intelligence},
  pages={217--227},
  year={2020},
  organization={PMLR}
}

@inproceedings{ahuja2023interventional,
  title={Interventional causal representation learning},
  author={Ahuja, Kartik and Mahajan, Divyat and Wang, Yixin and Bengio, Yoshua},
  booktitle={International conference on machine learning},
  pages={372--407},
  year={2023},
  organization={PMLR}
}

@article{young2014image,
  title={From image descriptions to visual denotations: New similarity metrics for semantic inference over event descriptions},
  author={Young, Peter and Lai, Alice and Hodosh, Micah and Hockenmaier, Julia},
  journal={Transactions of the Association for Computational Linguistics},
  volume={2},
  pages={67--78},
  year={2014},
  publisher={MIT Press One Rogers Street, Cambridge, MA 02142-1209, USA journals-info~…}
}

@article{bengio2013estimating,
  title={Estimating or propagating gradients through stochastic neurons for conditional computation},
  author={Bengio, Yoshua and L{\'e}onard, Nicholas and Courville, Aaron},
  journal={arXiv preprint arXiv:1308.3432},
  year={2013}
}

@article{podell2023sdxl,
  title={Sdxl: Improving latent diffusion models for high-resolution image synthesis},
  author={Podell, Dustin and English, Zion and Lacey, Kyle and Blattmann, Andreas and Dockhorn, Tim and M{\"u}ller, Jonas and Penna, Joe and Rombach, Robin},
  journal={arXiv preprint arXiv:2307.01952},
  year={2023}
}

@article{zhang2024clip,
  title={CLIP-MoE: Towards Building Mixture of Experts for CLIP with Diversified Multiplet Upcycling},
  author={Zhang, Jihai and Qu, Xiaoye and Zhu, Tong and Cheng, Yu},
  journal={arXiv preprint arXiv:2409.19291},
  year={2024}
}

@article{huang2024llm2clip,
  title={LLM2CLIP: Powerful Language Model Unlock Richer Visual Representation},
  author={Huang, Weiquan and Wu, Aoqi and Yang, Yifan and Luo, Xufang and Yang, Yuqing and Hu, Liang and Dai, Qi and Dai, Xiyang and Chen, Dongdong and Luo, Chong and others},
  journal={arXiv preprint arXiv:2411.04997},
  year={2024}
}

@article{bengio2000neural,
  title={A neural probabilistic language model},
  author={Bengio, Yoshua and Ducharme, R{\'e}jean and Vincent, Pascal},
  journal={Advances in neural information processing systems},
  volume={13},
  year={2000}
}

@article{mikolov2013distributed,
  title={Distributed representations of words and phrases and their compositionality},
  author={Mikolov, Tomas and Sutskever, Ilya and Chen, Kai and Corrado, Greg S and Dean, Jeff},
  journal={Advances in neural information processing systems},
  volume={26},
  year={2013}
}

@article{lavoie2024modeling,
  title={Modeling caption diversity in contrastive vision-language pretraining},
  author={Lavoie, Samuel and Kirichenko, Polina and Ibrahim, Mark and Assran, Mahmoud and Wilson, Andrew Gordon and Courville, Aaron and Ballas, Nicolas},
  journal={arXiv preprint arXiv:2405.00740},
  year={2024}
}
}

\clearpage
\setcounter{page}{1}
\maketitlesupplementary

\section{Proof for Theorem}


\begin{proof}

  Our proof consists of three steps.
  At \textbf{Step 1}, we show that the alignment term $\La $ in the learning objective \eqref{eq:contrastive_objective} enables encoders $ \ftt $ and $ \fii $ to identify the partitions $ \zt $ and $ \zi $. That is, the estimates $ \hzt $ and $ \hzi $ are equivalent to the true variables $ \zt $ and $ \zi $ up to invertible maps, respectively.

  At \textbf{Step 2}, we show the identifiability of the learned mask $ \hat{\mm} $ by proving the estimate $ \hat{\mm} $ is an invertible function of the true mask $ \mm $ with the sparsity, free from the influence of $ \zi $.

  At \textbf{Step 3}, we combine the intermediary results in \textbf{Step 1} and \textbf{Step 2} to show the block-wise identifiability of the latent variable $ \zi $.

  We define invertible maps: $ \rtt:= \ftt \circ \gtt $ and $\rii := \fii \circ \gii $, and denote the index set $ C: = \{ 1, \dots, d_{\zi} \} $. 

  \paragraph{Step 1.}
  We show by contradiction that for each mask $ \mm \in \cM $, the estimated partition $ \hzt:= [\rtt (\zt, \et)]_{C}  $ is an invertible function of the true partition $ \zt $ (i.e., $ \mm \odot \zi $) (i.e., $ \et $ doesn't influence $\hzt$).

  Following the generating process \eqref{eq:data_generating_process}, we express the estimated variable as: $ \hzi: = [\rii( \zi, \ei )]_{C} $.

  The alignment term $\La$ in \eqref{eq:contrastive_objective} enforces the invariance:
  \begin{align} \label{eq:invariance_z}
    \hzt &= \hat{\mm} \odot \hzi \nonumber \\
    &\implies \nonumber \\
    [\rtt (\zt, \et)]_{C} &= \hat{\mm} \odot [\rii( \zi, \ei )]_{C} \nonumber \\ 
    &\implies \nonumber \\
    [\rtt ( \mm \odot \zi, \et)]_{C} &= \hat{\mm} \odot [\rii( \zi, \ei )]_{C}.
  \end{align}

  Suppose that the exogenous variable $ \et $ influenced $\hzt$.
  Then there would exist a fixed point $ ( \zi^{*}, \et^{*} ) $ and indices $ (i^{*}, j^{*}) $ such that $ \frac{ \partial [\hzt]_{i^{*}} }{ \partial [\et]_{j^{*}} } ( \zi^{*}, \et^{*} ) \neq 0 $.
  Without loss of generality, suppose that $ \frac{ \partial [\hzt]_{i^{*}} }{ \partial [\et]_{j^{*}} } ( \zi^{*}, \et^{*} ) > 0 $.
  The smoothness of $ \rtt $ implies that this partial derivative would remain positive over a small neighborhood $ \cN := \{ \zi^{*} \} \times ( [\et^{*}]_{j} - \eta, [\et^{*}]_{j} + \eta) $, where we indicate the complement of the index $ j^{*} $ with $ -j^{*} $ and the radius with $ \eta >0 $.
  Therefore, $ [\hzt]_{i^{*}} $ would grow monotonically over the interval $\cN$.
  Nevertheless, the right-hand side of \eqref{eq:invariance_z} is not a function of $ \et $ and thus would remain constant over $ \cN $, leading to a contradiction.
  Thus, we have shown that $ \hzt $ is not a function of $\et$.
  Therefore, the estimate $ \hzt $ can be expressed as an invertible function $ \hzt $ of the true partition $ \zt $ (i.e., $ \mm \odot \zi $): $ \hzt = \htt (\zt) $.
  An identical argument applies to the latent variable $ \zi $ and gives rise to the invertible function $ \hii $ such that $ \hzi = \hii (\zi) $.   


  \paragraph{Step 2.}
  We start with developing basic equalities as a consequence of the true data-generating process \eqref{eq:data_generating_process} and the estimation objective \eqref{eq:contrastive_objective}.

  For any $(\zi, \mm)$, we can express the estimate $ \hzt $ according to the data-generating process \eqref{eq:data_generating_process}:
  \begin{align} \label{eq:hzt_dgp}
    \hzt = \htt ( \zt ) \underbrace{=}_{\eqref{eq:data_generating_process}} \htt ( \zi \odot \mm ),
  \end{align}

  Alternatively, the objective \eqref{eq:contrastive_objective} implies that
  \begin{align} \label{eq:hzt_obj}
    \hzt \underbrace{=}_{\La} \hat{\mm} \odot \hzi = \hat{\mm} \odot \hii ( \zi ),
  \end{align}
  where $ \hat{\mm} $ is a function of $ (\zi, \mm) $ (i.e., $\vt$).

  It follows from \eqref{eq:hzt_dgp} and \eqref{eq:hzt_obj} that:
  \begin{align} \label{eq:hzt_eq}
    \htt ( \zi \odot \mm ) = \hat{\mm} \odot \hii ( \zi ).
  \end{align}
  
  Now, we show that if $ \vt^{(1)} $ and $ \vt^{(2)} $ originate from any two $ \zi^{(1)} $ and $ \zi^{(2)} $ (potentially identical) under one specific mask $\mm$, i.e., they belong to one group $ \mT(\mm) $, their masks in the estimated process should also be identical $ \hat{\mm}^{(1)} = \hat{\mm}^{(2)} $.

  Suppose that they were assigned to distinct masks in the estimated process.
  That is, the function $ \hat{\mm} ( \zi \odot \mm ) $ takes distinct values on $ \zi^{(1)} $ and $ \zi^{(2)} $ under a fixed $\mm$: $ \hat{\mm} ( \zi^{(1)} \odot \mm ) \neq \hat{\mm} ( \zi^{(2)} \odot \mm ) $.
  Then, we can find a boundary point $ \tzi^{(1)} $, such that for any arbitrary ball centered at it contains points that are assigned with a distinct group: $ \forall \epsilon > 0, \exists \tzi^{(2)} \in \cB( \tzi^{(1)}, \epsilon ) $ such that $ \hat{\mm} ( \tzi^{(1)} \odot \mm ) \neq \hat{\mm} ( \tzi^{(2)} \odot \mm ) $.
  Therefore, we can find a sequence $ \{ \tzi^{(2, i)} \}_{n}$ that converges to $ \tzi^{(1)} $: $ \tzi^{(2, n)} \to \tzi^{(1)} $.
  The continuity of the function $ \htt ( \zi \odot \mm ) $ in $ \zi $ implies that $  \htt ( \tzi^{(2, n)} \odot \mm ) \to  \htt ( \tzi^{(1)} \odot \mm )$, which together with \eqref{eq:hzt_eq} indicates that
  \begin{align} \label{eq:contradiction}
    \hat{\mm}^{(2)} \odot \hii ( \zi^{(2, i)} ) \to \hat{\mm}^{(1)} \odot \hii ( \zi^{(1)} ).
  \end{align}

  Note that we have $ \hat{\mm}^{(1)} \neq \hat{\mm}^{(2)} $. 
  If for dimension $i$, $[\hat{\mm}^{(1)} ]_{i} = 1 $ and $ [ \hat{\mm}^{(2)} ]_{i} = 0 $,
  \eqref{eq:contradiction} implies that $ \hii ( \zi^{(1)} ) = 0 $.
  In the other case $[\hat{\mm}^{(1)} ]_{i} = 0 $ and $ [ \hat{\mm}^{(2)} ]_{i} = 1 $,
  \eqref{eq:contradiction} implies that  $\hii ( \zi^{(2, i)} ) \to 0$, which indicates that $ \hii ( \zi^{(1)} ) = 0 $.

  Therefore, we have shown that for any $ \vt^{(1)} $ and $ \vt^{(2)} $ belonging to the same group $ \mT(\mm) $, their assigned estimated masks $ \hat{\mm}^{(1)} $ and $ \hat{\mm}^{(2)} $ would only differ on zero-valued coordinates of $ \hii ( \zi^{(1)} ) $ and $\hii ( \zi^{(2)} )$.
  The regularization $ \Ls $ in \eqref{eq:contrastive_objective} selects the sparest masks $ \hat{\mm}^{(1)} $ and $ \hat{\mm}^{(2)} $ and sets these coordinates in the corresponding estimated masks $ \hat{\mm}^{(1)} $ and $ \hat{\mm}^{(2)} $ to zero, as this does not affect the minimization of $\La$.
  Therefore, we have shown that the estimated masks are equal $ \hat{\mm}^{(1)} = \hat{\mm}^{(2)} $, if the original masks agree $ \mm^{(1)} = \mm^{(2)} $.

  Because the estimated group count and the true ground count are identical, any distinct original masks $ \mm^{(1)} \neq \mm^{(2)} $ should give rise to distinct estimated masks $ \hat{\mm}^{(1)} \neq \hat{\mm}^{(2)} $.

  Therefore, we have shown that the equality of estimated masks $ \hat{\mm}^{(1)} = \hat{\mm}^{(2)} $ is equivalent to that of the original masks $ \mm^{(1)} = \mm^{(2)} $.

  Due to the invertibility and smoothness of $\htt$ (in particular, its injectivity and continuity) and $ \hii $, they preserve the dimensionality of their domains as subsets of the Euclidean space $ \R^{d(\zi)} $. 
  Thus, \eqref{eq:hzt_eq} implies that the true and the estimated masks have the same number of non-zeros values $ \norm{\mm}_{0} = \norm{\hat{\mm}}_{0}$. 
  Thus, we can automatically learn the grouping information and the correct mask sparsity in an unsupervised manner.

  \paragraph{Step 3.}
  It follows from \eqref{eq:hzt_eq} that the estimated block $ \hat{\mm} \odot \hzi $ is an invertible map of the true latent variable block $ \mm \odot \zi $, where the estimated and the true masks are related via a bijection.
  Identification of the unions of blocks follows directly from the identification of all constituent blocks.
  Regarding intersections , for any two intersecting masks $ \mm^{(1)} $ and $ \mm^{(2)} $ in the support $ \cM $ such that $ \mm^{(1)} \odot \mm^{(2)} \neq \0 $, the intersection block associated with their product $ \mm^{(1)} \odot \mm^{(2)} $ is also identifiable.
  We denote the index sets on which two masks are non-zero as $ \cB^{(1)} := \{ i \in d(\zi): [\mm^{(1)}]_{i} = 1\} $ and $ \cB^{(2)} := \{ i \in d(\zi): [\mm^{(2)}]_{i} = 1\} $.

  Given the relation \eqref{eq:hzt_eq}, we have that:
  \begin{align*}
    \htt ( \zi \odot \mm^{(1)} ) &= \hat{\mm}^{(1)} \odot \hii ( \zi ) \\
    \htt ( \zi \odot \mm^{(2)} ) &= \hat{\mm}^{(2)} \odot \hii ( \zi ).
  \end{align*}
  This implies that the estimates $ \hzt $ are aligned on components indexed with $ \cB^{(1,2)} := \cB^{(1)} \cap \cB^{(2)}$:
  \begin{align}
    \begin{split}
      &[\hii ( \zi )]_{ \cB^{(1,2)} } \\ 
      &=[\htt ( \zi \odot \mm^{(1)} )]_{\cB^{(1,2)}} = [\htt ( \zi \odot \mm^{(2)} )]_{ \cB^{(1,2)} }.
    \end{split}
  \end{align}
  The identical reasoning in \textbf{Step 1} implies that the intersection $ [\hzi]_{ \cB^{(1,2)} } $ is an invertible function of the true partition $[\zi]_{ \cB^{(1,2)} }$.
  Therefore, we can identify the intersection of any subsets of the masks with the same argument.

  This concludes our proof.

\end{proof}

\section{More Text-to-Image Generation Results}
In this section, we present more text-to-image generation results with different CLIP models.

\begin{figure*}
    \centering
    \begin{tabular}{ccc}
        \hline
        \multicolumn{3}{p{0.9\textwidth}}{
        The red panda sits on a thick bamboo branch, its fluffy tail curled around the stalk for balance. The panda's round face wears a gentle expression, with bright eyes that sparkle like morning dew. Its russet fur glows warmly in the filtered sunlight, making it look like a living plushie. The panda's little black paws grip a piece of bamboo like a precious toy, while its round ears stand alert against the green background. Behind the panda, stalks of rainbow-colored bamboo create a magical forest scene, their leaves casting soft shadows. \textcolor{red}{\textbf{Small butterflies rest on nearby leaves, their wings matching the panda's red-gold fur}}. A collection of fallen leaves creates a natural confetti effect around the branch where the panda sits. The panda's distinctive facial markings look extra striking against its fluffy white cheeks, like nature's perfect face paint. Soft clouds visible through the bamboo leaves frame this peaceful scene, where the adorable creature enjoys its cozy perch.
        } \\ \\ \hline
        CLIP & LongCLIP & SmartCLIP (Ours)\\ 
        \includegraphics[scale=0.1]{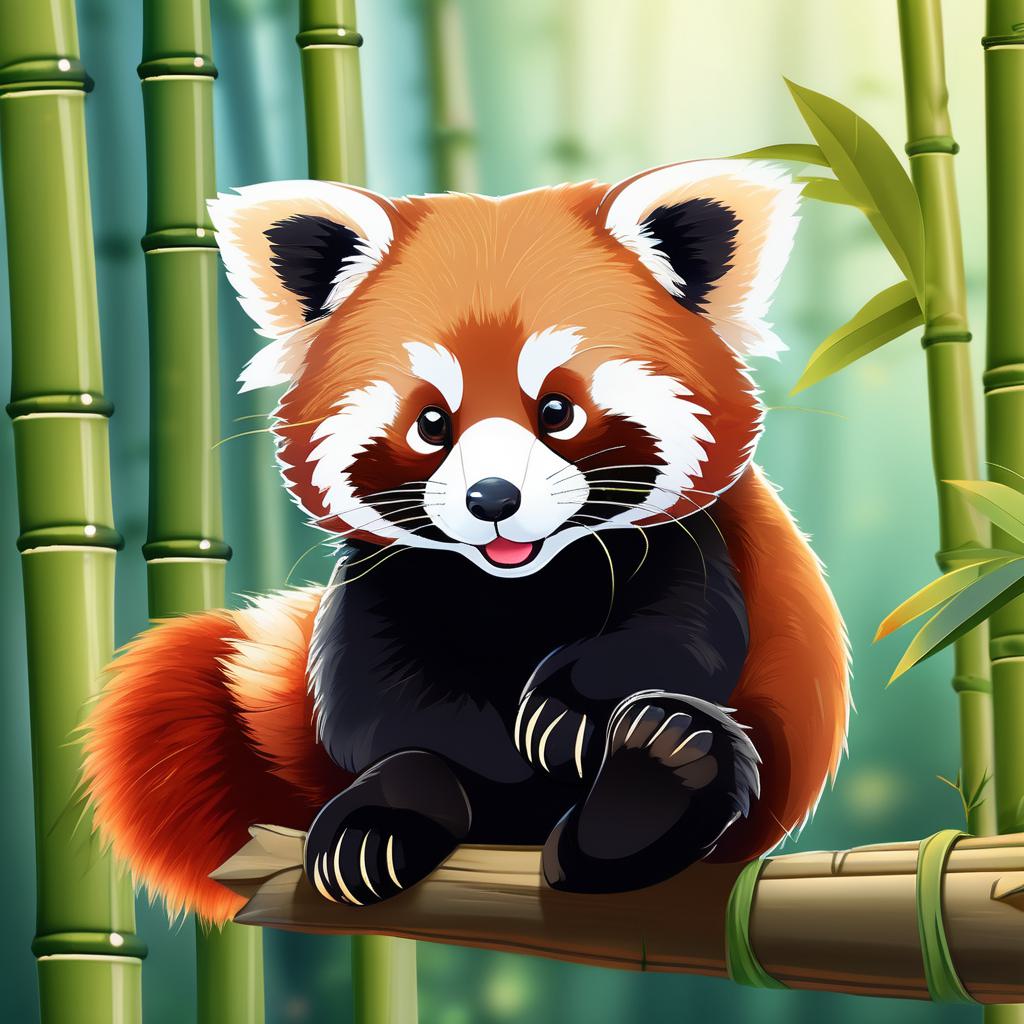}
        &
        \includegraphics[scale=0.1]{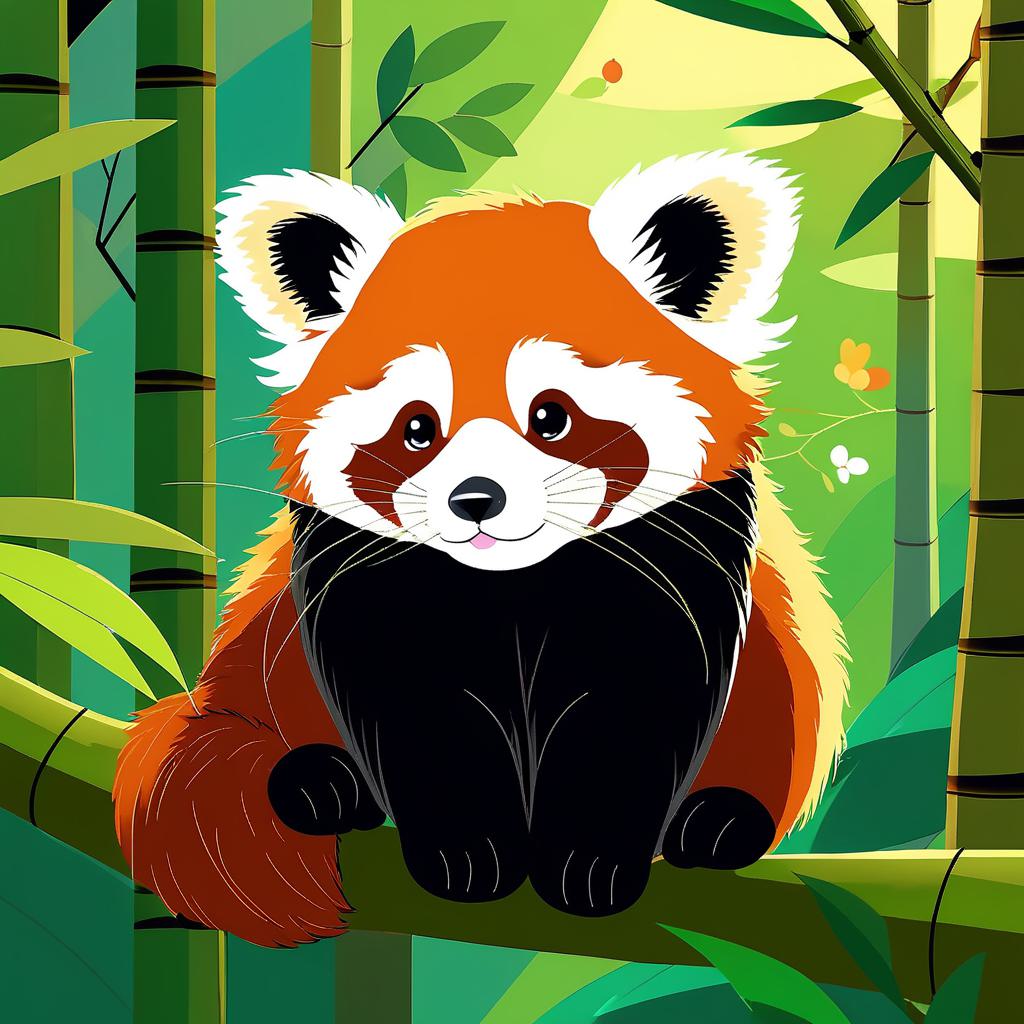}
        &
        \includegraphics[scale=0.1]{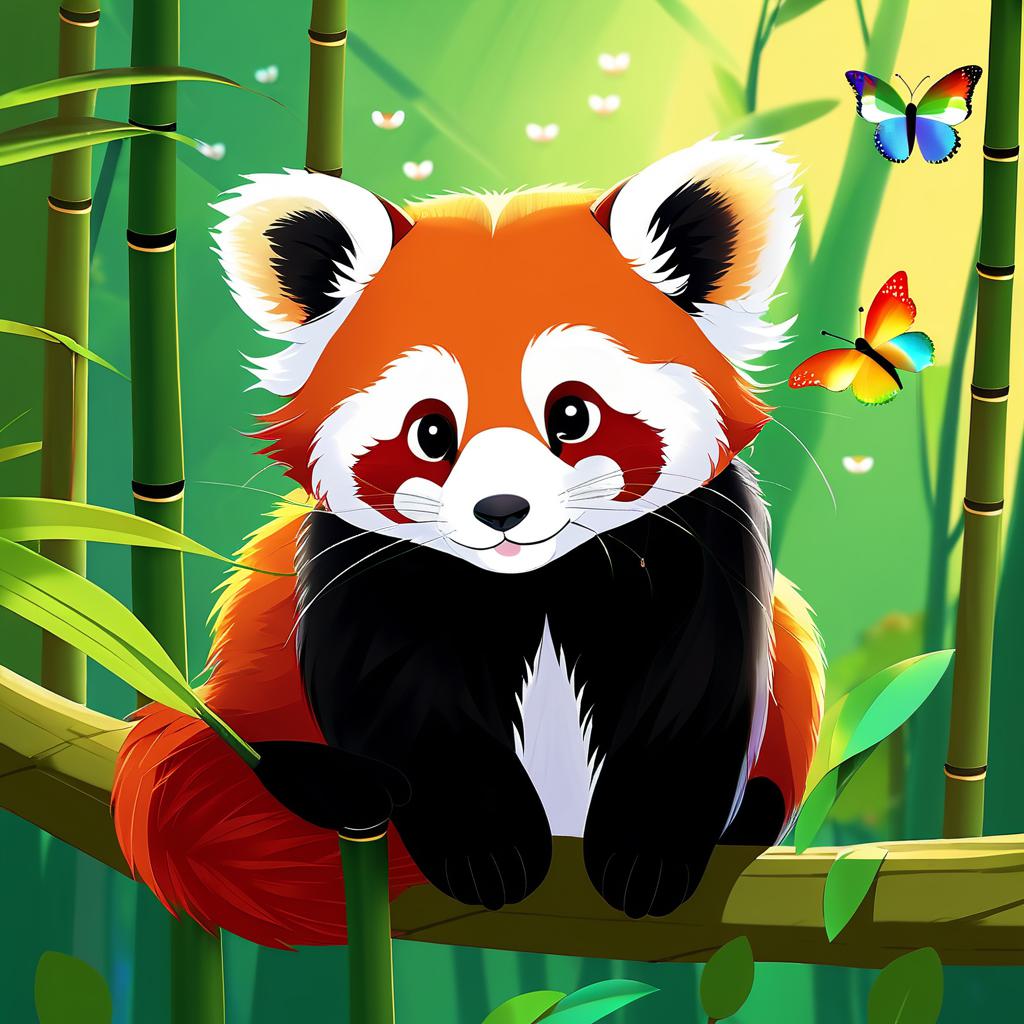} \\ \hline \\
        
         \multicolumn{3}{p{0.9\textwidth}}{
       A smiling Corgi sits confidently on what appears to be a stroller or cart with handlebar grips, looking absolutely delighted with its tongue hanging out. The dog's cream and ginger fur glows in the urban lighting, while its pointed ears stand at attention. The fashionable pup is transformed into a true urban explorer with its eye-catching accessories. The sleek sunglasses frame makes a bold statement against its furry face, giving the Corgi a cool, confident look. \textcolor{red}{\textbf{The stylish fedora hat sits perfectly between its ears}}, completing the sophisticated ensemble. Those trendy sunglasses and the jaunty fedora hat create an irresistible tourist charm. The fashionable eyewear and classic hat combination transforms this ordinary Corgi into a social media-worthy city slicker. URBAN SETTING: \textcolor{red}{\textbf{The scene takes place in what appears to be Times Square or a busy city intersection, with taxis, bright signs, and pedestrians visible in the background.}}
        } \\ \\ \hline
        CLIP & LongCLIP & SmartCLIP (Ours)\\ 
        \includegraphics[scale=0.1]{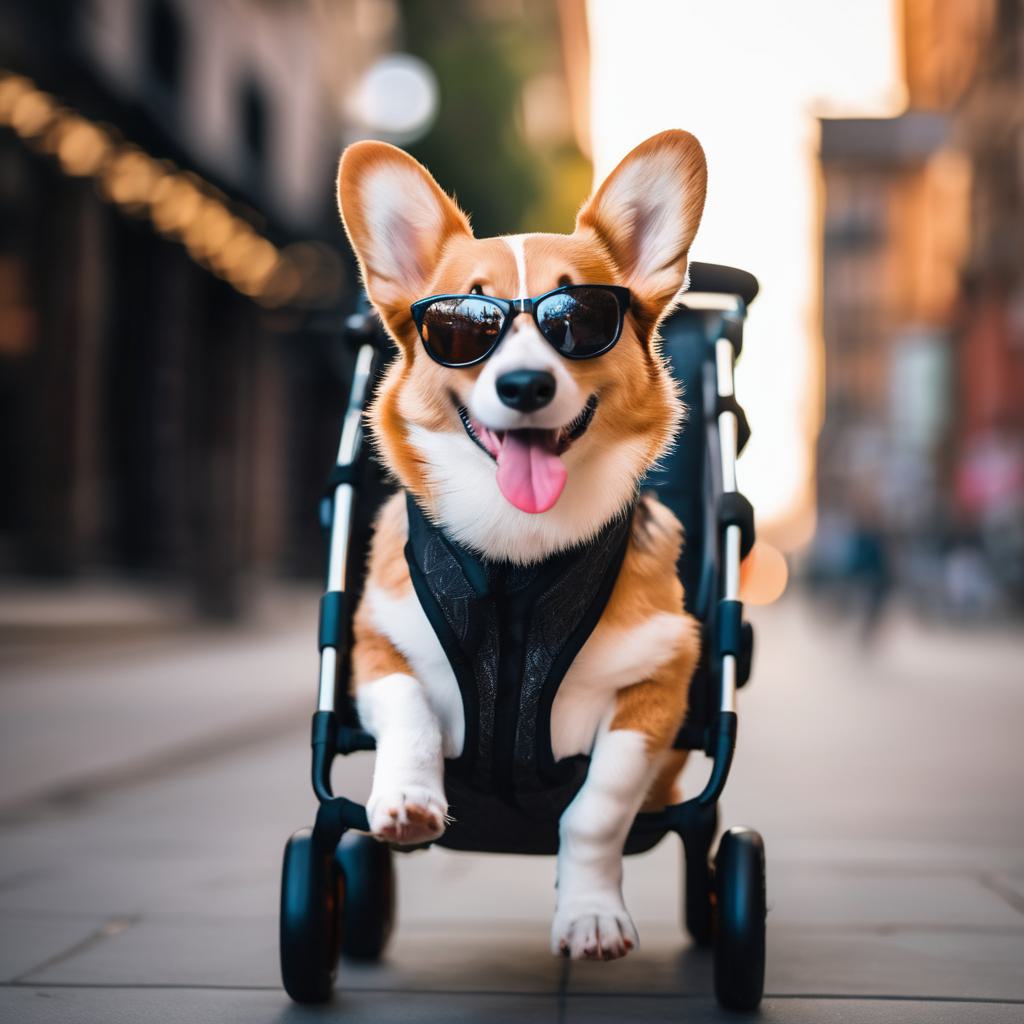}
        &
        \includegraphics[scale=0.1]{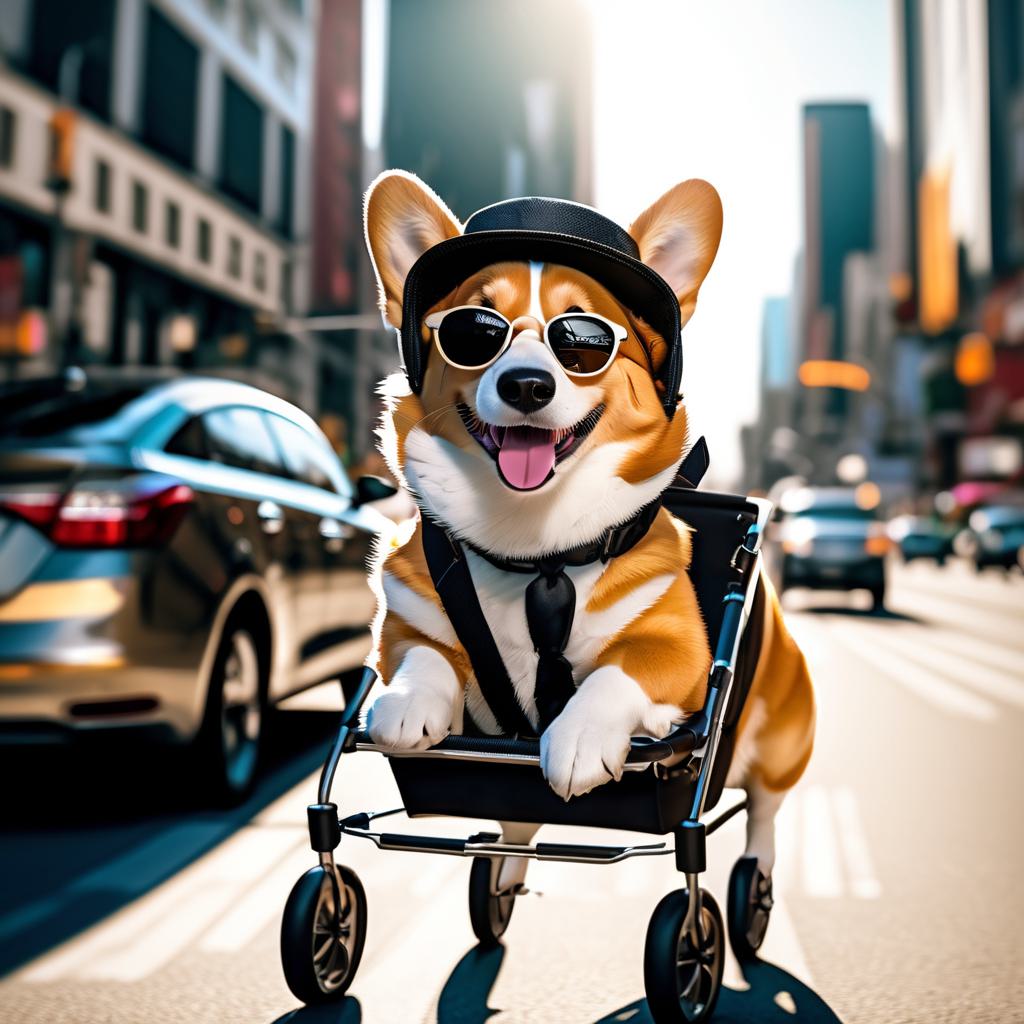}
        &
        \includegraphics[scale=0.1]{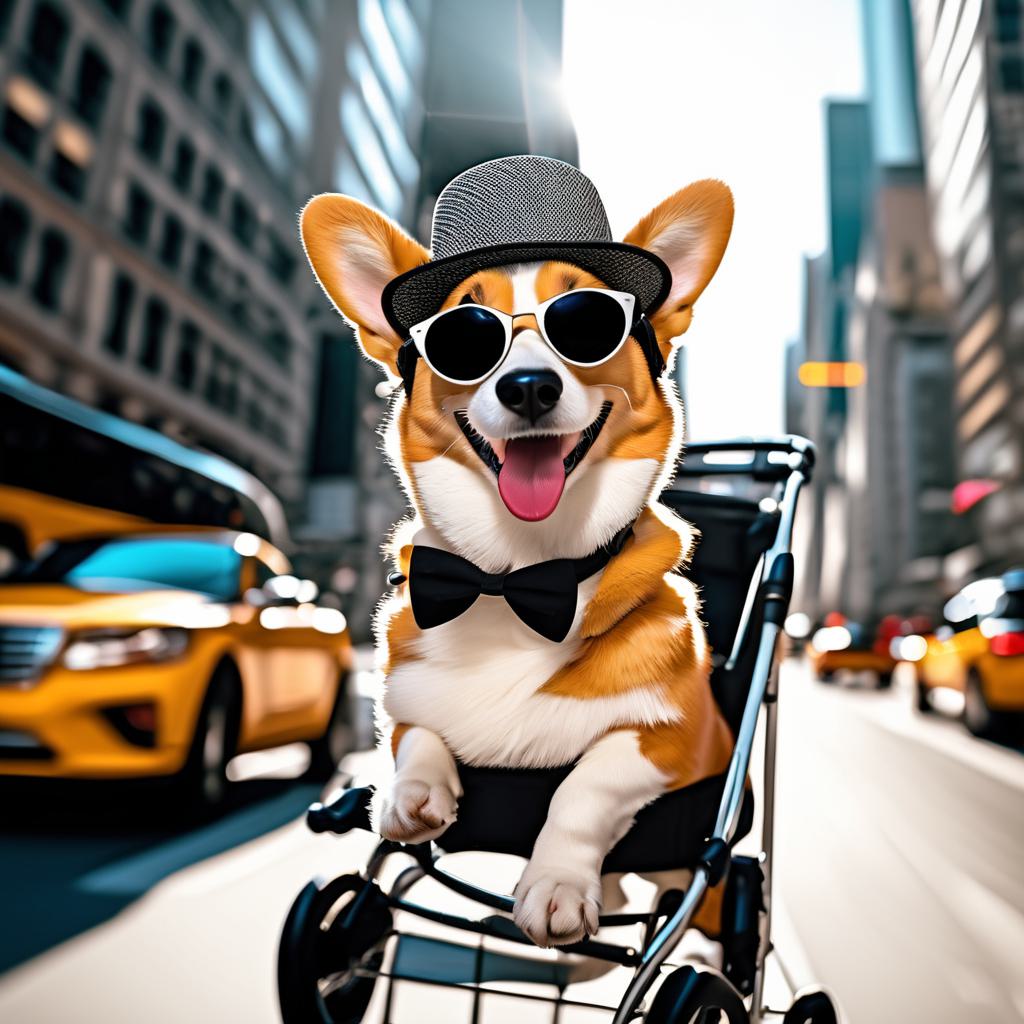} \\
        
        \hline
    \end{tabular}
    \caption{Long text-to-image generation comparions.}
    \label{fig:t2i1}
\end{figure*}

\begin{figure*}
    \centering
    \begin{tabular}{ccc}
        \hline
        \multicolumn{3}{p{0.9\textwidth}}{
       A whimsical Renaissance-style portrait showcasing a raccoon with its distinctive grey and white masked face against a dark dramatic background. The painting is presented in an ornate baroque golden frame with elaborate scrollwork and flourishes. ROYAL ATTIRE: Perched majestically upon the raccoon's head sits a \textcolor{red}{gleaming golden crown adorned with jewels and intricate details}, symbolizing royal authority. The regal golden crown catches the light, emphasizing its metalwork and precious stones. The raccoon wears a \textcolor{red}{\textbf{luxurious red robe with detailed gold embroidery}}, perfectly complementing the magnificent crown that marks its royal status. The portrait's composition draws attention to the resplendent crown, which features delicate spires and ornamental elements typical of royal regalia. The entire scene is unified by the interplay between the golden crown, the rich red garments, and the elaborate frame, creating a masterful parody where the crown serves as the ultimate symbol of the raccoon's noble bearing.
        } \\ \\ \hline
        CLIP & LongCLIP & SmartCLIP (Ours)\\ 
        \includegraphics[scale=0.1]{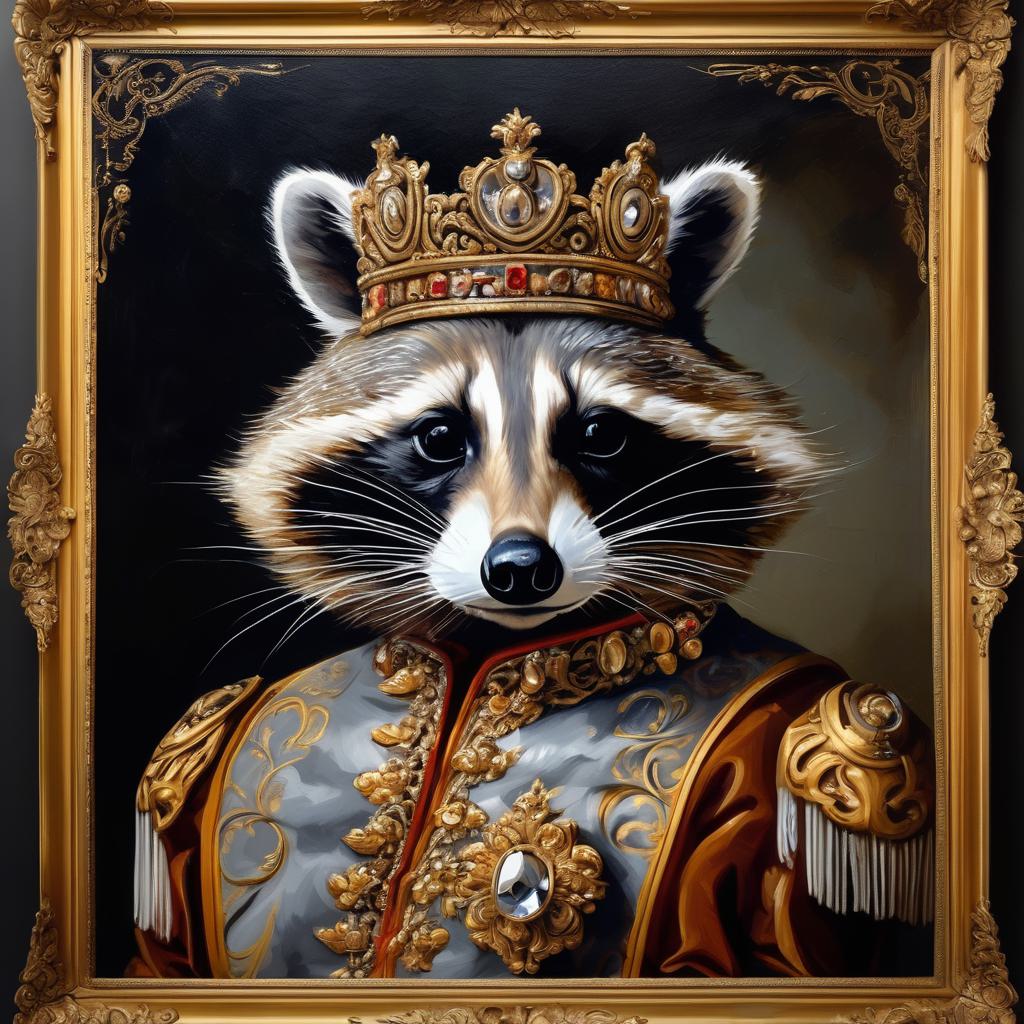}
        &
        \includegraphics[scale=0.1]{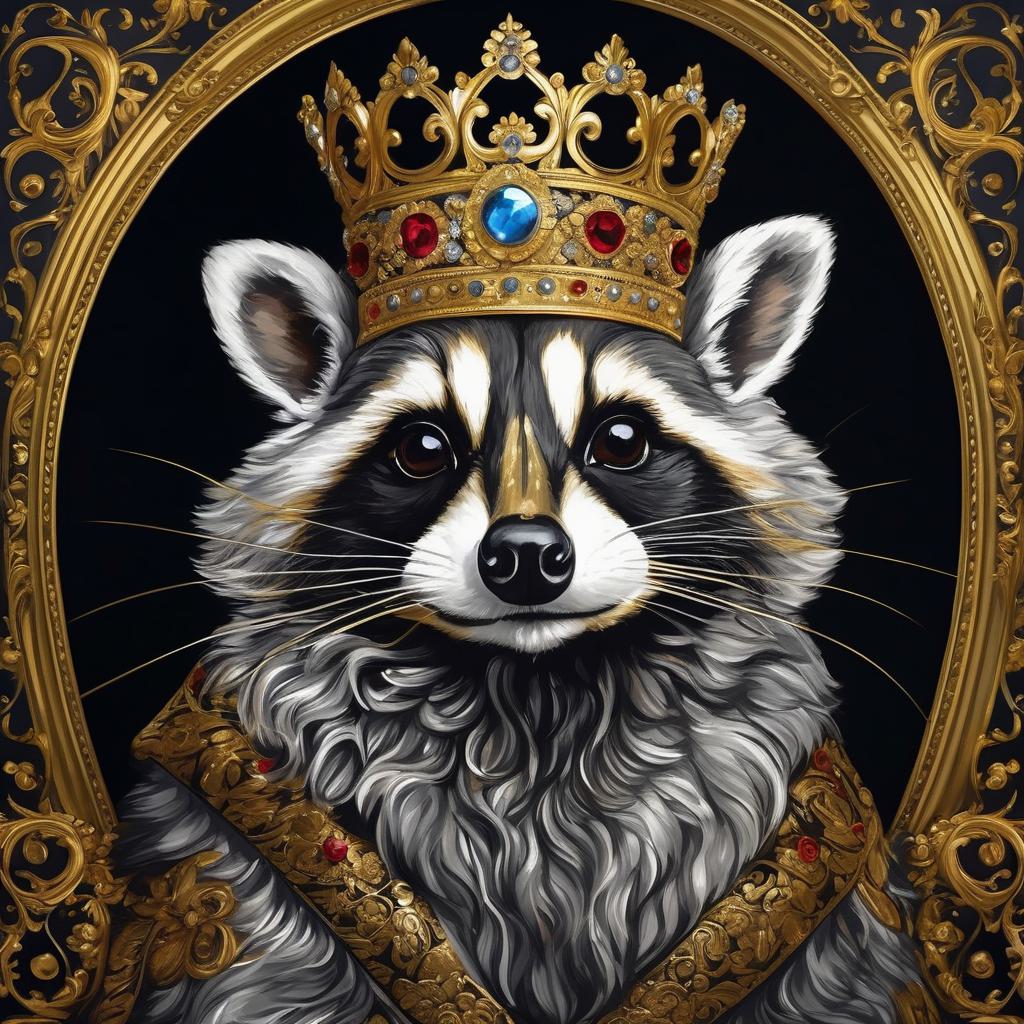}
        &
        \includegraphics[scale=0.1]{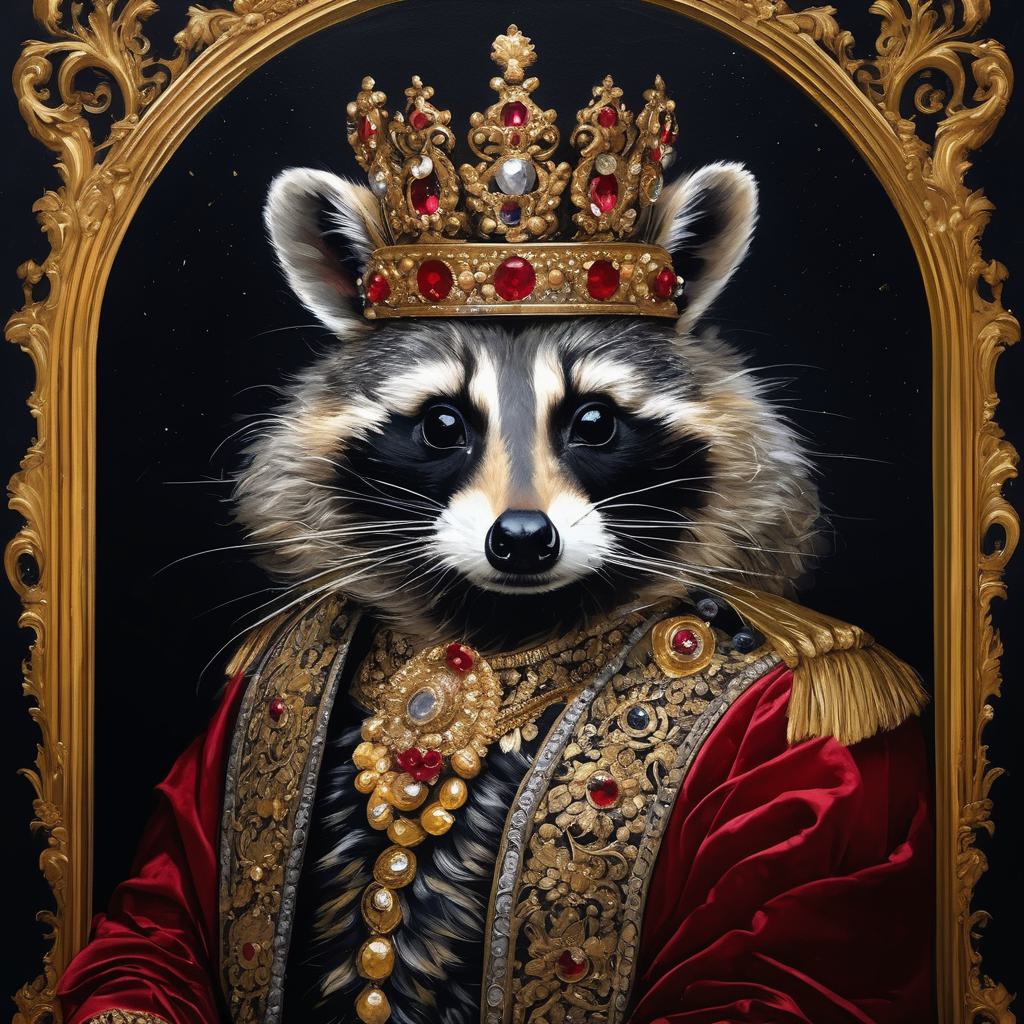} \\ \hline \\
        
         \multicolumn{3}{p{0.9\textwidth}}{
       A regal Dachshund with a rich reddish-brown coat stands atop a wooden surface, photographed in beautiful mood lighting. The dog's distinctive long body, short legs, and alert expression are captured in sharp detail against a rustic wooden background. The scene exudes warmth and autumn atmosphere. DECORATIVE ELEMENTS: Below the dog, arranged artistically, sits a glowing vintage lantern casting golden light, accompanied by pumpkins
 . The antique lantern's warm illumination bathes the scene in cozy light, while \textcolor{red}{\textbf{the pumpkins}} add seasonal charm. The composition is completed with scattered logs and a blue plaid blanket, where the lit lantern creates dramatic shadows and the autumn pumpkins rest naturally. The vintage lantern, rustic pumpkins, and thoughtfully arranged props transform this pet portrait into a quintessential fall scene.
        } \\ \\ \hline
        CLIP & LongCLIP & SmartCLIP (Ours)\\ 
        \includegraphics[scale=0.1]{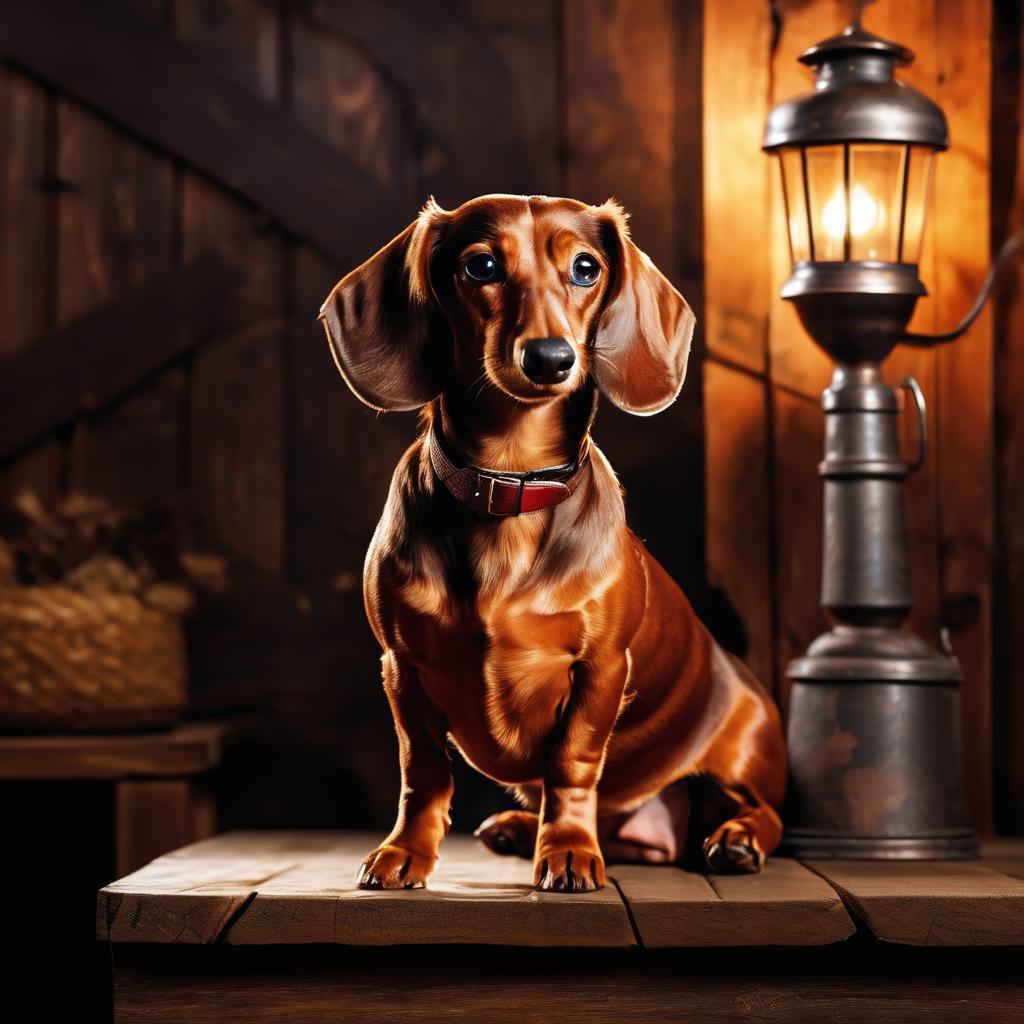}
        &
        \includegraphics[scale=0.1]{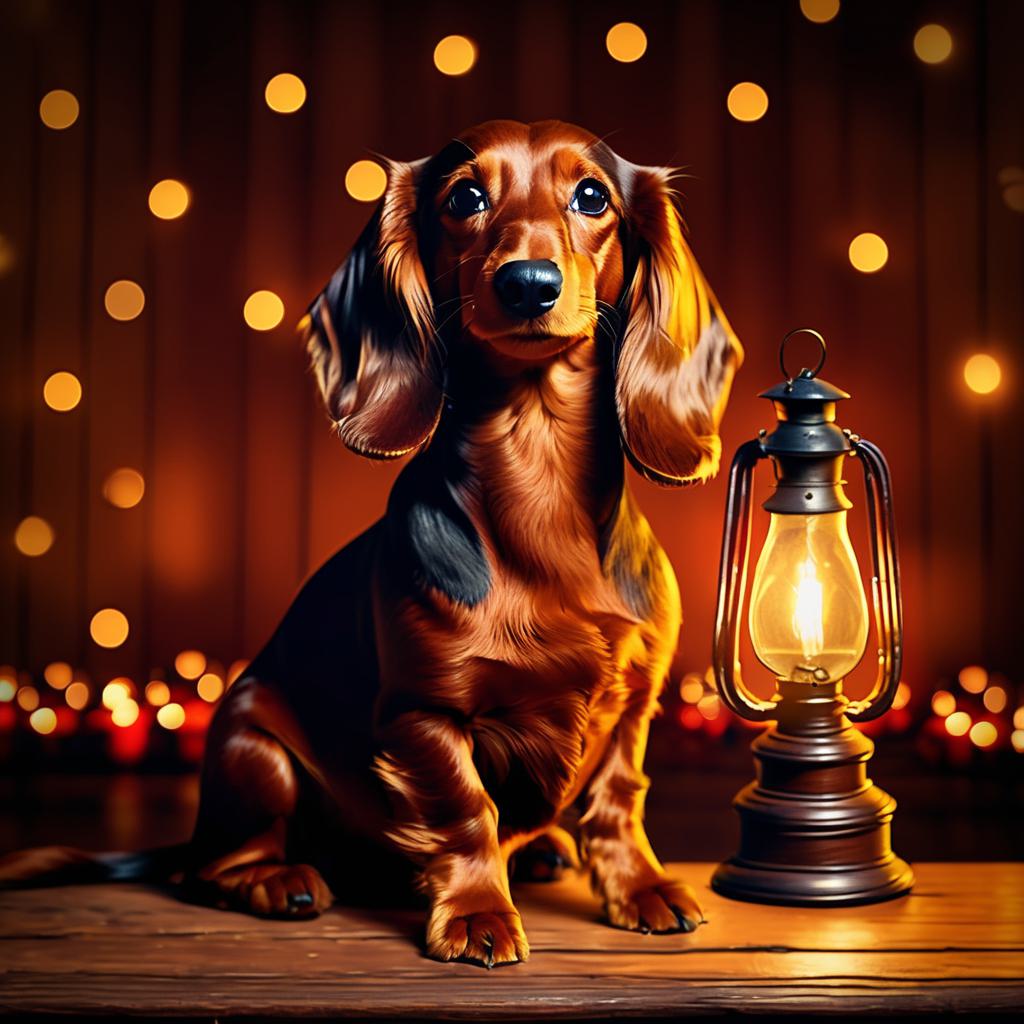}
        &
        \includegraphics[scale=0.1]{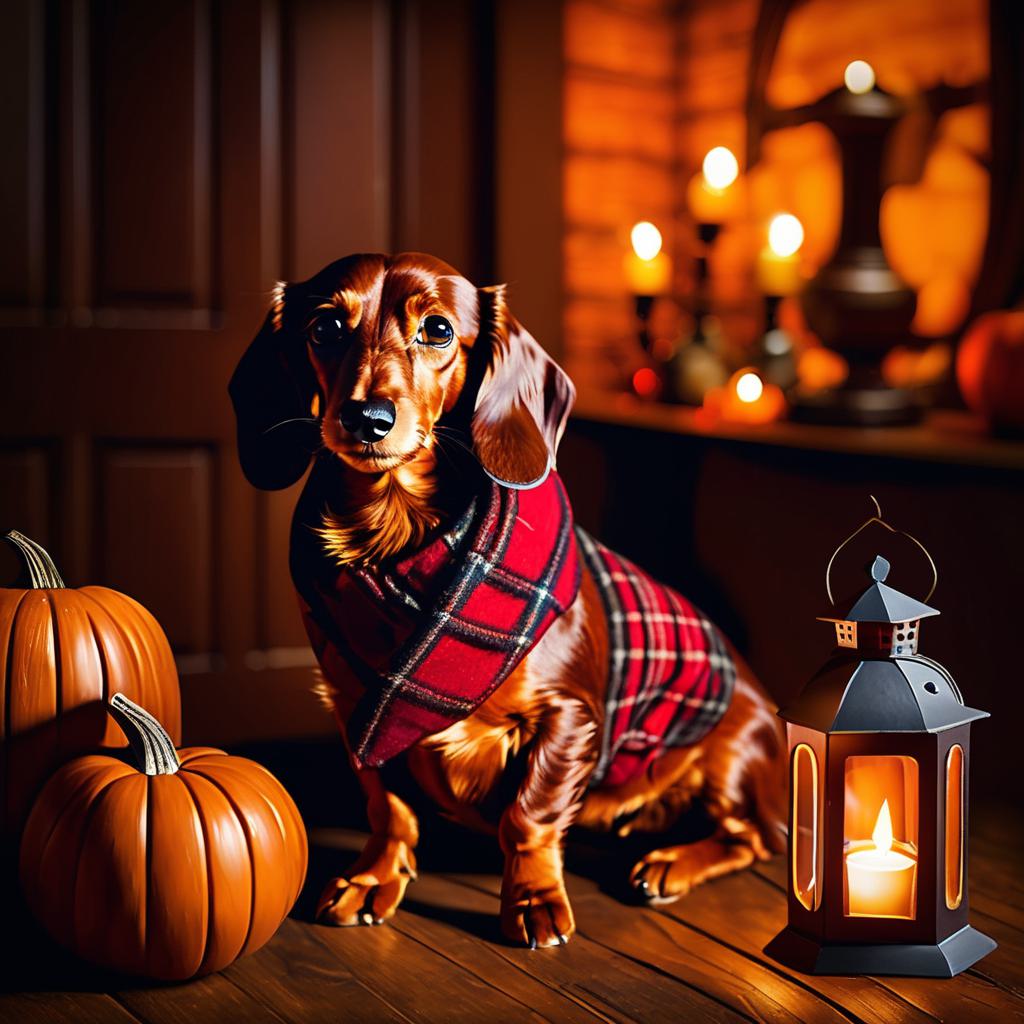} \\
        
        \hline
    \end{tabular}
    \caption{Long text-to-image generation comparions.}
    \label{fig:t2i2}
\end{figure*}

\end{document}